\documentclass[journal]{IEEEtran}

\usepackage{xcolor,soul,framed} 
\usepackage{pifont}
\colorlet{shadecolor}{yellow}
\usepackage{graphicx}
\graphicspath{{../pdf/}{../jpeg/}}
\DeclareGraphicsExtensions{.pdf,.jpeg,.png}

\usepackage[cmex10]{amsmath}
\usepackage{array}
\usepackage{mdwmath}
\usepackage[utf8]{inputenc}
\usepackage{textgreek}
\usepackage{mdwtab}
\usepackage{eqparbox}
\usepackage{url}
\usepackage{booktabs}
\usepackage{amssymb}
\usepackage{multirow}
\usepackage{dblfloatfix}
\usepackage{booktabs}
\usepackage{siunitx}
\usepackage{adjustbox}
\usepackage{etoolbox}
\usepackage{tabularx}
\newcommand{\TableBodyFont}{\footnotesize}
\AtBeginEnvironment{tabular}{\TableBodyFont}
\AtBeginEnvironment{tabular*}{\TableBodyFont}
\AtBeginEnvironment{tabularx}{\TableBodyFont}
\newcolumntype{L}[1]{>{\raggedright\arraybackslash}p{#1}}
\newcolumntype{Y}{>{\centering\arraybackslash}X}
\begin{document}
\bstctlcite{IEEEexample:BSTcontrol}
    \title{Task-Specified Active Metrological Inspection with Measurement-Steered VLA Manipulation and Deterministic Evidence Gating}
\author{
Zhiling Chen$^1$,
Jingzhan Ge$^1$,
Ruimin Chen$^1$,
Matthew P. Castanier$^2$,
David Gorsich$^2$,
Farhad Imani$^1$

\thanks{$^1$ Zhiling Chen, Jingzhan Ge, Ruimin Chen, and Farhad Imani are with the University of Connecticut, Storrs, CT, USA.\\
$^2$ David Gorsich and Matthew P. Castanier are with the U.S. Army DEVCOM Ground Vehicle Systems Center (GVSC), Warren, MI, USA.\\
\\
DISTRIBUTION STATEMENT A. Approved for public release; distribution is unlimited. OPSEC11090}%
}



\maketitle

\begin{abstract}
High-mix low-volume (HMLV) manufacturing requires inspection systems to adapt to changing parts, specifications, and work orders without repeated task-specific programming. Existing inspection automation typically assumes predefined sensing sequences, while general purpose robot agents optimize task completion rather than the completeness and validity of metrological evidence. We formulate task-specified active metrological inspection and propose From Requirements to Admissible Metrological Evidence (FRAME), a hierarchical dual-arm framework that converts an inspection instruction and structured specification into traceable conformance evidence. FRAME coordinates learned manipulation with calibrated laser profilometry: a task manager grounds and schedules requirements, active surface correspondence verifies physical-to-specification localization, and evidence memory tracks measurement provenance, admissibility, and coverage. Learned components may propose inspection targets and physical access actions, but deterministic datum-grounded measurement, admissibility checks, coverage auditing, and conformance evaluation prevent incomplete or unverified evidence from authorizing PASS. A series of physical experiments shows that FRAME achieves higher end-to-end inspection reliability, fewer false accepts, and shorter task completion time.

\end{abstract}


%
\IEEEpeerreviewmaketitle


\section{Introduction}

\begin{figure*}[!t]
\centering
\includegraphics[width=\textwidth]{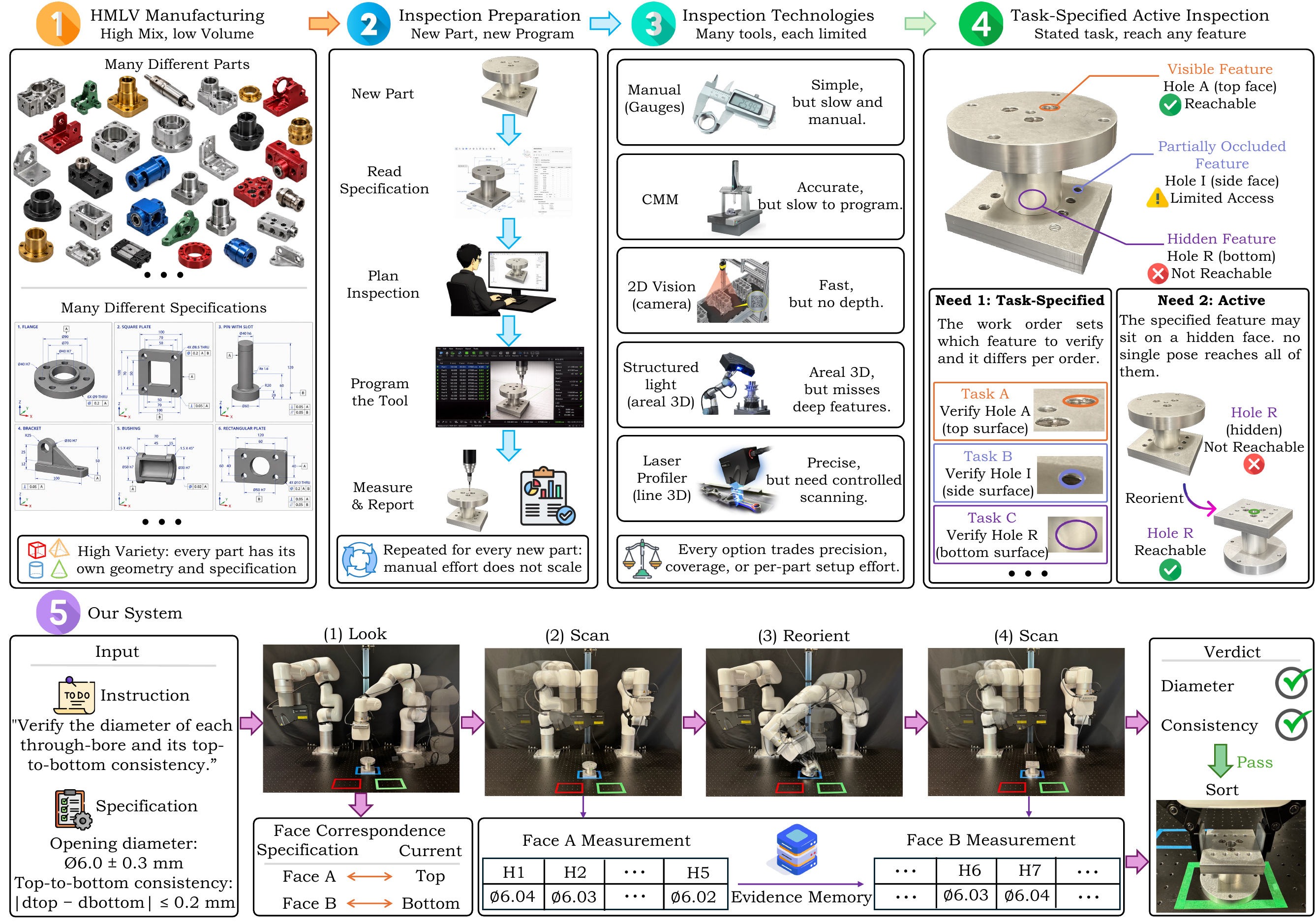}
\caption{Motivation and overview of task-specified active inspection for HMLV manufacturing. High part and specification variability makes conventional inspection preparation costly, while existing inspection technologies trade off accuracy, coverage, and per-part setup effort. A stated inspection task may require features distributed across visible, partially occluded, and hidden surfaces, motivating both task-specified measurement and active reorientation. Our system takes a natural-language instruction and structured specification, grounds the required features to the current part pose, alternates \textsc{look}, \textsc{scan}, and \textsc{reorient} to acquire verified evidence, produces a deterministic conformance verdict, and invokes \textsc{sort} for an authorized binary outcome.}
\label{fig:introduction}
\end{figure*}

\IEEEPARstart{H}{igh-mix} low-volume (HMLV) manufacturing requires quality inspection to adapt to frequent changes in parts, specifications, and inspection intents. Unlike mass production, where the cost of a fixed inspection program can be amortized over many identical units, HMLV production repeatedly introduces small-batch variants with different geometries and inspection semantics, as illustrated in Fig.~\ref{fig:introduction}~(1). This regime resembles \emph{variational automation}, in which a robot persistently handles task instances with nontrivial variations in object geometry and pose, rather than \emph{fixed automation}, which repeatedly executes the same motions~\cite{chen2026gap}. In this work, we consider its metrological counterpart: the physical inspection setup remains fixed, while the inspection task, part geometry, object category, and initial pose may vary across jobs. Product evolution can change features, datum definitions, nominal dimensions, and tolerances, while different work orders for the same part may request a full conformance report, a particular feature, or a functional group of requirements. Each change can therefore restart a human-intensive workflow of reading the specification, planning the inspection, programming and validating the tool, acquiring measurements, and preparing the report. The recurring preparation process in Fig.~\ref{fig:introduction}~(2), rather than measurement time alone, becomes a central obstacle to scalable HMLV inspection. What is needed is a system that determines from the current instruction and specification what evidence is required instead of executing a fixed part-specific inspection plan.

Existing inspection technologies automate individual sensing and measurement operations, but not this complete specification-to-evidence workflow. Manual gauges are flexible but slow and operator-dependent; coordinate measuring machines provide accurate and traceable measurements but generally rely on a prepared setup and validated program; two-dimensional vision is fast but does not directly provide the required three-dimensional geometry; area three-dimensional sensing can recover surfaces but remains limited by occlusion and difficult feature access; and a laser line profiler provides precise localized geometry only when it follows a controlled scan trajectory. Consequently, introducing a new part still requires an engineer to interpret its specification and plan the sequence summarized in Fig.~\ref{fig:introduction}~(2). As Fig.~\ref{fig:introduction}~(3) emphasizes, every available tool trades precision, coverage, or per-part setup effort. Flexible inspection therefore requires more than an accurate sensor: it must decide what to inspect, how to make the required regions measurable, and when the acquired evidence is sufficient.

Robot foundation models have recently made rapid progress in turning language and vision into generalizable physical actions. Vision-language-action (VLA) models scale language-conditioned visuomotor control across tasks and embodiments~\cite{brohan2023rt,kim2025openvla,black2026pi0visionlanguageactionflowmodel}, while world-action models (WAMs) jointly predict future visual states and actions to model physical dynamics and improve transfer to unseen motions and environments~\cite{ye2026world, yuan2026fast, ma2026dit4dit}. However, stronger foundation policies do not by themselves yield reliable long-horizon systems: existing VLAs remain vulnerable to skill-chaining, subtask-dependency, and progress-tracking errors that compound over multi-step execution~\cite{fan2025long,liu2026palm}, while WAMs can still produce inaccurate imagined futures whose predicted outcomes may diverge from the physical states realized during execution~\cite{ye2026world, wang2026trust}. A complementary lesson has emerged from digital agents, where \emph{harness engineering} improves practical capability by surrounding a foundation model with structured tool interfaces, persistent state, execution feedback, validation, and opportunities to revise failed decisions~\cite{yang2024swe,zhong2026ai}. Early efforts brought language-model reasoning into robot execution through generated control programs and composable perception-and-control representations~\cite{liang2023code,huang2023voxposer}; more recent robotics work has begun to adopt the fuller harness principle by exposing learned policies as bounded skills and orchestrating them with high-level planners, analytic primitives, memory, verification, and retry mechanisms~\cite{zhang2026harness,lee2026harness, thea2026}. This shift improves composition and recovery, but metrological inspection imposes a stricter system-level requirement: the harness must ensure not only that a physical action succeeds, but also that every required characteristic is supported by complete, correctly localized, and admissible evidence before a \textsc{pass} verdict can be issued.

We therefore study \emph{task-specified active metrological inspection}. Given an inspection instruction, a structured specification, and an initial observation of a physical part, the robot must infer the task-induced requirement scope, actively acquire complete admissible evidence for that scope, and return a traceable inspection report and operational verdict. The setting is \emph{task-specified} because different work orders over the same part may require different features or functional relations, and it is \emph{active} because the required regions may be visible, partially occluded, or inaccessible from the initial configuration, as shown in Fig.~\ref{fig:introduction}~(4). The robot must consequently choose among observation, reorientation, scanning, recovery, and final sorting rather than execute a fixed sensing sequence. This problem has three safety-relevant failure sources: the inferred scope may omit a required requirement, the system may admit evidence from a physical region that does not correspond to the required specification region, or the metrological process may classify a truly nonconforming requirement as conforming. The central challenge is therefore not merely exposing a hidden feature, but preventing an error at any of these stages from silently supporting \textsc{pass}.

We propose From Requirements to Admissible Metrological Evidence (FRAME), a hierarchical framework for task-specified active metrological inspection that coordinates semantic task management, active physical access, deterministic metrology, and evidence-gated conformance. The execution flow is illustrated in Fig.~\ref{fig:introduction}~(5). A task manager grounds the instruction to identifiers in the structured specification, maintains a surface-correspondence table between physical regions and specification regions, and selects the next action from the uncovered requirements in evidence memory. Calibrated \textsc{look} actions establish or verify the current correspondence; a VLA-based manipulation expert performs contact-rich reorientation and final sorting; and a measurement expert performs calibrated line-profile acquisition, datum-grounded measurement, and requirement-specific evidence-admissibility checks. Failed correspondence, acquisition, or admissibility checks trigger another observation, reorientation, or scan. Evidence memory is not a store of model-generated conclusions: it is a traceable ledger of measurements, their physical origins, admissibility outcomes, and requirement coverage. Only after every requirement in the inferred scope has been supported by admissible evidence does a deterministic, specification-driven conformance evaluator produce the final verdict. We evaluate FRAME on task-specified active metrological inspection tasks spanning four controlled workpiece families, five inspection-task families, and additional CNC cases. Empowered by verified surface correspondence and evidence-gated conformance, FRAME demonstrates higher end-to-end reliability and shorter task-completion time while producing fewer false accepts than fixed exhaustive inspection and unconstrained VLM control in physical experiments.

The main contributions of this article are as follows.

\begin{enumerate}

    \item We formulate \emph{task-specified active metrological inspection} in terms of an instruction-induced requirement scope, active surface access, admissible evidence coverage, and specification-driven conformance, and derive an auditable decomposition of its false-accept risk.

    \item We develop a hierarchical active-inspection framework that connects a task manager, calibrated viewpoint control, a VLA-based manipulation expert, a deterministic measurement expert, a surface-correspondence table, evidence memory, and a specification-driven conformance evaluator.
    
    \item We introduce datum-grounded measurement operators, requirement-specific evidence-admissibility predicates, post-reorientation verification, active recovery, and coverage-gated verdict generation so that a learned proposal cannot directly authorize \textsc{pass}.
    
    \item We construct a controlled benchmark spanning four workpiece families and five inspection-task families, and evaluate FRAME on a physical dual-arm platform through end-to-end comparisons, metrological validation, and component-level analyses of requirement-scope grounding, surface correspondence, and diagnostic recovery.

\end{enumerate}


\section{Related Work} \label{sec: related work}

\subsection {Industrial Inspection and Metrology Automation}

Industrial inspection automation spans dimensional metrology, active geometric acquisition, and appearance-based defect detection. In dimensional metrology, CAD-based coordinate measuring machine planners generate collision-free probing sequences, while robot-mounted laser-scanner methods optimize trajectory overlap and measurement uncertainty~\cite{lin2001cad,phan2018path,vlaeyen2022uncertainty}. These methods automate a predefined measurement plan but commonly assume a product model, designated features, or feature-specific scan strategies. Active acquisition instead selects informative viewpoints or trajectories to improve geometric coverage, with recent work extending this formulation toward instruction-conditioned high-precision surface scanning~\cite{chen2024gennbv,chen2025scanbot}. A complementary line detects and localizes visual defects from nominal image data and increasingly supports language interaction~\cite{bergmann2019mvtec,roth2022towards,gu2024anomalygpt}. However, reconstruction methods do not determine requirement-level conformance, while anomaly detectors do not produce datum-grounded quantitative measurements. Our framework instead derives the required inspection scope from the current instruction and specification, actively exposes the corresponding surfaces, and admits only localized, metrologically valid measurements into coverage-gated conformance.

\subsection{Robot Foundation Models}

Robot foundation models provide increasingly generalizable physical manipulation across tasks, objects, and embodiments. Vision-language-action models and generalist robot policies scale language-conditioned visuomotor control, while world-action models complement action generation with future-state prediction~\cite{brohan2022rt,bousmalis2023robocat,team2024octo,ye2026world}. These capabilities are well suited to changing the physical accessibility of inspection surfaces under task, geometry, and pose variation, but active metrological inspection cannot be closed by manipulation alone. It requires a planner that coordinates learned physical actions with a calibrated inspection sensor and uses structured measurement feedback to determine whether an acquisition should be repeated, another surface should be exposed, a remaining requirement should be measured, or the inspection can terminate. This coordination becomes particularly important over long horizons, where skill-chaining, subtask-dependency, and progress-tracking errors can accumulate~\cite{fan2025long,liu2026palm}. Our framework therefore uses a foundation policy to realize contact-rich reorientation, while the planner converts measurement validity, evidence admissibility, requirement coverage, and recovery diagnostics into the next sensing or manipulation subgoal, thereby allowing precision-sensor feedback to steer physical execution without asking the learned policy itself to interpret or certify metrological results.

\subsection {Agentic Robot Planning}

Agentic robot systems extend foundation models from direct action generation to planning and orchestration over modular physical skills. Language-model planners map open-ended instructions to admissible or affordance-grounded actions; closed-loop agents further incorporate environmental feedback, interleave reasoning with tool use, and retain experience across trials~\cite{huang2022language,brohan2023can,huang2022inner,yao2022react,shinn2023reflexion}. Physical-agent systems bring these ideas to real robots by integrating perception and control modules, expressing goals as optimizable constraints, orchestrating learned primitives with recovery, and placing runtime structure around foundation policies~\cite{liu2024demonstrating,huang2024rekep,li2026roboclaw,chen2026gap,zhang2026harness}. These approaches show that performance depends not only on the underlying policy but also on how skills are exposed, sequenced, monitored, verified, and retried. However, their feedback loops are generally organized around action progress or task completion, whereas active metrological inspection must use structured measurement outcomes to determine what physical action is needed next. Our framework therefore orchestrates learned manipulation and calibrated sensing in a shared loop: measurement validity, evidence admissibility, remaining requirement coverage, and recovery diagnostics determine whether the planner issues \textsc{look}, \textsc{scan}, \textsc{reorient}, \textsc{reacquire}, or termination.

\section{Problem Formulation}
\label{sec: problem formulation}

We study \emph{task-specified active metrological inspection}. Given an inspection instruction $I$, a structured specification $S$, and an initial observation $o_0$ of a physical part $P$, the robot must acquire the required evidence and return a traceable report with either a supported conformance decision or a fail-safe unresolved outcome. Unlike a fixed inspection program, $I$ selects the requirements for the current episode, while $S$ defines their interpretation and evaluation.

\subsection{Task-Specified Inspection Scope}
\label{sec:problem-scope}

We represent the structured specification as
\begin{equation}
S = (\mathcal{C},\mathcal{F}, \mathcal{R}_{\mathrm{all}}, \mathcal{D}, \mathcal{A}, \Psi),
\label{eq:specification}
\end{equation}
where $\mathcal C$ is a finite vocabulary of specification-declared task concepts and their expansions to leaf requirements; $\mathcal F$ contains surface-region identifiers; $\mathcal R_{\mathrm{all}}$ is the universe of inspectable leaf requirements; and $\mathcal D$, $\mathcal A$, and $\Psi$ define datum semantics, evidence-admissibility predicates, and conformance logic. Regions are general inspectable patches, with \emph{face} used only for naturally face-like regions. Leaf requirements may represent geometric characteristics, surface conditions, or inter-feature and inter-region relations. The specification declares their dependencies, nominal values, tolerances, datums, topology, admissibility rules, and decision logic, but not which concepts a particular production decision requests. Let
\begin{equation}
\mathcal C^{*}=\rho^{*}(I,S)\subseteq\mathcal C,
\qquad
R^{*}=\operatorname{Expand}_{S}(\mathcal C^{*})
\subseteq \mathcal{R}_{\mathrm{all}}
\label{eq:required-scope}
\end{equation}
denote the ground-truth task concepts and induced leaf-requirement scope. $\operatorname{Expand}_{S}$ follows only dependencies declared in $S$; for example, an assembly-readiness concept may expand to requirements for a locating interface, mounting pattern, and supporting surface. Full conformance is the special case $R^{*}=\mathcal{R}_{\mathrm{all}}$. Task specification therefore changes the required scope, not its completeness: every requirement in $R^{*}$ still needs admissible evidence.
At execution time, the robot estimates $\widehat{\mathcal C}$ from $I$ and $S$ and deterministically obtains $\widehat R=\operatorname{Expand}_{S}(\widehat{\mathcal C})$. An omission $R^{*}\nsubseteq\widehat R$ is safety relevant, whereas extra requirements increase cost without directly causing a false accept. $S$ is a machine-interpretable representation of drawing or semantic PMI inspection semantics, using the subset of ASME Y14.5 datum and GD\&T constructs supported by single-sided line-profile acquisition~\cite{asme_y145_2018}. 


\subsection{Active Inspection and Surface Correspondence}
\label{sec:problem-actions}

The specification names target regions symbolically, whereas the robot must locate physical regions that provide the required evidence. Under the inspection-cell presentation protocol, a part arrives in a finite set $\mathcal X_c$ of support-stable scanning configurations for object class $c$. Each configuration encodes support contact, a discrete orientation class, and the robot--sensor arrangement, and determines the physically scannable regions $\mathcal F^{\mathrm{scan}}(x)$. Membership requires a collision-free, kinematically feasible sensor trajectory with valid standoff, incidence, field of view, and measurement range, while providing sufficient coverage of the target region and any datum support required by the measurement; visual exposure alone is insufficient.
The initial configuration $x_0\in\mathcal X_c$ is unknown, but a generic locating surface or upstream handler restricts continuous translation and rotation to a calibrated capture range. Unconstrained orientations and unstable placements are outside our scope. Let $\mathcal{F}^{\mathrm{phys}}$ denote the physical surface regions of the part, and let
\begin{equation}
\mu^{*}: \mathcal{F}^{\mathrm{phys}} \rightarrow \mathcal{F}\cup\{\bot\}
\label{eq:surface-correspondence}
\end{equation}

denote the ground-truth physical-to-specification correspondence, with $\bot$ marking regions absent from the active specification. The currently scannable specification regions are therefore $\mu^{*}(\mathcal F^{\mathrm{scan}}(x_t))\setminus\{\bot\}$. Before scan evidence can support conformance, the robot must infer the discrete association $\widehat\mu_t$.

The robot interacts with the part through the action set
\begin{equation}
a_t \in \{\textsc{look},\textsc{reorient},\textsc{scan},\textsc{finish/sort}\}.
\label{eq:action-set}
\end{equation}
\textsc{Look} identifies $x_t$ and checks \eqref{eq:surface-correspondence}; \textsc{reorient} changes the support-stable configuration through a flip, roll, turn, or regrasp; and \textsc{scan} acquires data from $\mathcal F^{\mathrm{scan}}(x_t)$. \textsc{Finish/sort} is permitted only after report and decision generation. Inspection is therefore active because required evidence may span configurations, while an incorrect reorientation must not silently corrupt the decision.

\subsection{Evidence and Coverage}
\label{sec:problem-evidence}

The robot accumulates an evidence memory $\mathcal{M}_t=\{e_1,\ldots,e_t\}$. A feature-level evidence record is represented as
\begin{equation}
e=(\widehat r,\widehat q,b,\ell),
\label{eq:evidence-record}
\end{equation}
where $\widehat r$ is the associated requirement, $\widehat q$ the measured quantity, $b\in\{0,1\}$ its admissibility, and $\ell$ its traceability metadata and diagnostics.
$b$ is not a conformance decision: $b=1$ only when the evidence satisfies the requirement-specific conditions in $\mathcal A$, including feature coverage, boundary validity, point density, and absence of truncation. Records with $b=0$ cannot support PASS. $\ell$ retains the scan identifier, physical and specification regions, acquisition time, and failed conditions. 

Let $e\models r$ denote that an evidence record contains the measurements required by $r$ and originates from the physical surface region or regions correctly associated with the specification regions required by $r$. We define complete evidence coverage as
\begin{equation}
\operatorname{Cov}(\mathcal{M}_t,R)=1
\Longleftrightarrow
\forall r\in R,\ \exists e\in\mathcal{M}_t:
b(e)=1\ \wedge\ e\models r.
\label{eq:evidence-coverage}
\end{equation}
This ground-truth definition uses the latent correspondence $\mu^{*}$; runtime coverage instead uses $\widehat\mu_t$. PASS requires complete estimated coverage of $\widehat R$, even after one nonconformance is found. Missing or inadmissible evidence drives further acquisition or, after recovery is exhausted, an unresolved outcome rather than measured part nonconformance.

\subsection{Measurement-Steered Acquisition and Operational Outcomes}
\label{sec:problem-steering}

Evidence acquisition follows deficits in $\mathcal M_t$ rather than a fixed sensing sequence. The uncovered requirements are
\begin{equation}
\mathcal U_t=
\left\{
r\in\widehat R\;\middle|\;
\nexists e\in\mathcal M_t:
b(e)=1\ \wedge\ e\models_{\widehat\mu_t} r
\right\},
\label{eq:uncovered-requirements}
\end{equation}
where $e\models_{\widehat\mu_t}r$ denotes runtime support under the estimated correspondence. Each scan produces a candidate record and diagnostic
\begin{equation}
\begin{aligned}
d_t&=\operatorname{Diagnose}(e_t,S),\\
d_t&\in\mathcal D_{\mathrm{diag}},\\
\mathcal D_{\mathrm{diag}}
&=\{\textsc{admitted},\textsc{reacquire},\\
&\hspace{1.5em}\textsc{change-access},\textsc{unresolved}\}.
\end{aligned}
\label{eq:measurement-diagnostic}
\end{equation}
These outcomes distinguish admitted evidence, local reacquisition, a required change of access, and exhausted recovery.

The subsequent action is selected from the current correspondence and evidence state,
\begin{equation}
a_{t+1}=\pi_{\mathrm{act}}
\!\left(I,S,o_{0:t},\widehat\mu_t,
\mathcal M_t,\mathcal U_t,d_t\right).
\label{eq:measurement-steered-action}
\end{equation}
The process is \emph{measurement-steered} when an evidence deficit or failed admissibility condition changes the next physical action. \textsc{Reacquire} adjusts acquisition in the current configuration, whereas \textsc{change-access} reorients the part to expose a missing feature, support, or datum and requires independent verification before scanning. Steering is evidence-driven rather than value-driven: an admitted out-of-tolerance result is retained as nonconformance, not reacquired. The formulation leaves $\pi_{\mathrm{act}}$ and the access executor unspecified.

For complete evidence, the evaluator applies the datum semantics, deterministic tests, and verdict rule $\Psi$ to produce report $\widehat E$ and conformance verdict
\begin{equation}
\widehat y_{\mathrm{conf}}
\in \{\mathrm{PASS},\mathrm{NO\mbox{-}PASS}\},
\label{eq:estimated-verdict}
\end{equation}
defined only under complete estimated coverage. The operational outcome is
\begin{equation}
\widehat y
\in
\{\mathrm{PASS},\mathrm{NO\mbox{-}PASS},\mathrm{UNRESOLVED}\}.
\label{eq:operational-outcome}
\end{equation}
\textsc{Pass} and \textsc{no-pass} are supported $\widehat y_{\mathrm{conf}}$ decisions; \textsc{unresolved} denotes exhausted recovery with incomplete evidence.

\subsection{Auditable False-Accept Risk}
\label{sec:problem-risk}

A false accept releases PASS when the task-induced ground-truth verdict is NO-PASS. Let $y^{*}$ be the result of applying $S$ to ground-truth evidence for $R^{*}$:
\begin{equation}
P_{\mathrm{FA}}=
\Pr\!\left[
\widehat y=\mathrm{PASS}
\ \land\
y^{*}=\mathrm{NO\mbox{-}PASS}
\right].
\label{eq:false-accept}
\end{equation}

Here, $P_{\mathrm{FA}}$ is the joint false-accept probability over
all episodes. Let
$\pi_{\mathrm{NP}}=\Pr(y^{*}=\mathrm{NO\mbox{-}PASS})$.
The corresponding conditional false-accept probability satisfies
\[
P_{\mathrm{FA}}
=
\pi_{\mathrm{NP}}
\Pr\!\left(
\widehat y=\mathrm{PASS}
\mid
y^{*}=\mathrm{NO\mbox{-}PASS}
\right).
\]

A false accept has three sources. First, the inferred scope may omit a requirement:
\begin{equation}
G=\{R^{*}\nsubseteq\widehat R\},
\qquad
\gamma=\Pr(G).
\label{eq:scope-error}
\end{equation}
Second, let $D_f$ denote the episode-level event that at least one wrong physical-to-specification association supporting required region $f$ is admitted during the episode, and let
$\delta=\max_f\Pr(D_f)$.
Third, let $A_r$ denote the episode-level event that, under correct scope and correspondence, a truly nonconforming requirement $r\in R^{*}$ is evaluated from admitted evidence and assigned PASS through erroneous evidence admission, numerical measurement, or conformance comparison. Let
$\alpha=\max_r\Pr(A_r)$. If only per-attempt error rates are available, all authorized attempts that can support a requirement during the episode must
be included in the corresponding union bound. Let
\begin{equation}
F_R=\bigcup_{r\in R^{*}}\mathcal{F}(r)
\label{eq:required-regions}
\end{equation}
be the required specification regions, where $\mathcal F(r)$ gives the regions on which $r$ depends. Because PASS requires complete evidence for $\widehat R$, a false accept must contain a scope omission, accepted correspondence error, or requirement-level metrological misclassification. Hence,
\begin{equation}
P_{\mathrm{FA}}
\leq
\gamma
+
\sum_{f\in F_R}\Pr(D_f)
+
\sum_{r\in R^{*}}\Pr(A_r)
\leq
\gamma+|F_R|\delta+|R^{*}|\alpha.
\label{eq:false-accept-bound}
\end{equation}

Equation~\eqref{eq:false-accept-bound} is an auditable risk contract that does not require an independence assumption, rather than an unconditional guarantee. Its terms are separately estimable from annotated scopes, correspondence ground truth, and traceable metrology ground truth, allowing the resulting bound to be compared with the observed end-to-end false-accept rate. Measurement steering governs recovery and efficiency, whereas the scope, correspondence, admissibility, coverage, and conformance gates determine whether the available evidence can support \textsc{pass}.


\section{Method}
\label{sec:method}

\subsection{System Overview}
\label{sec:method-overview}

\begin{figure*}[t]
\centering
\includegraphics[width=0.9\textwidth]{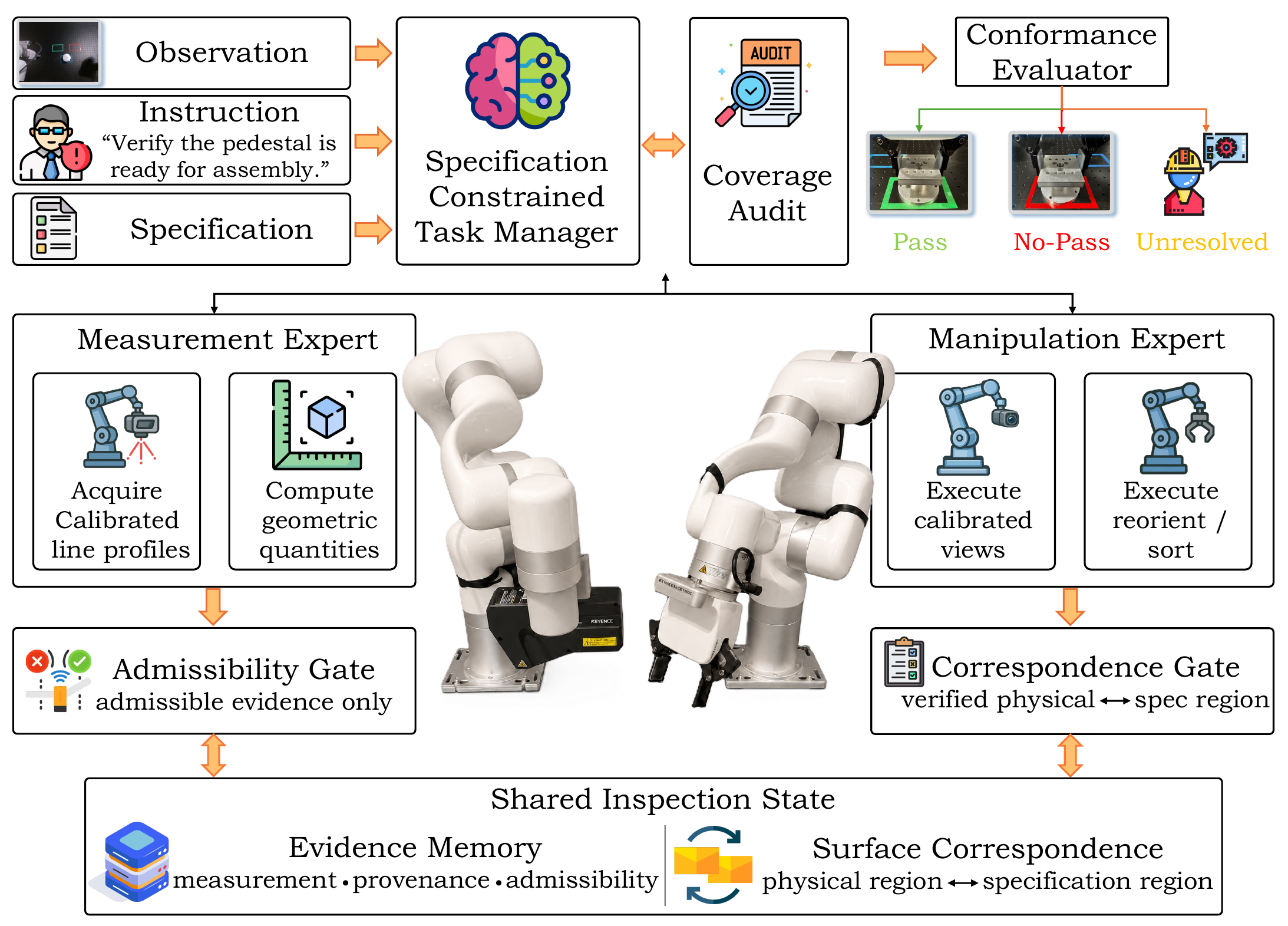}
\caption{Overview of our framework. The specification-constrained task manager coordinates a manipulation expert and a measurement expert through shared Surface Correspondence and Evidence Memory. The manipulation expert exposes three atomic skills: pre-calibrated \textsc{look}, VLA-based \textsc{reorient}, and VLA-based \textsc{sort}. Each reorientation is independently verified by \textsc{look} and the correspondence gate before scanning. The measurement expert converts calibrated profiles into admissible geometric evidence, while failed acquisitions trigger \textsc{reacquire}. Complete coverage enables deterministic conformance evaluation: a supported \textsc{pass} or \textsc{no-pass} verdict triggers \textsc{sort}, whereas exhausted execution or reacquisition budgets produce \textsc{unresolved} and manual review.}
\label{fig:model}
\end{figure*}

The framework closes semantic planning and physical interaction around deterministic evidence diagnostics. Fig.~\ref{fig:model} summarizes its task manager, two heterogeneous experts, shared inspection state, and deterministic evidence gates. The task manager grounds the instruction to specification-declared concepts, expands them into leaf requirements, and routes observation, measurement, and physical-access subtasks from the current correspondence and evidence state. The \emph{manipulation expert} exposes two inspection-time atomic skills: \textsc{look} executes a pre-calibrated camera-view action, while \textsc{reorient} uses a VLA policy to realize contact-rich changes of physical access. The same expert performs final sorting after a verdict is authorized. The \emph{measurement expert} converts calibrated laser profiles into feature-level geometric quantities and structured admissibility diagnostics. The components communicate through a surface correspondence table, which associates observed physical regions with task-specified regions, and an evidence memory, which stores admitted records and their acquisition provenance; the diagnostic status of every attempted acquisition is retained in the traceability log.

Given an instruction $I$, a structured specification $S$, and an initial observation $o_0$, the task manager first produces a specification-constrained scope $\widehat R$ and establishes the initial surface correspondence from calibrated camera observations. It then dispatches pose-grounded subtasks according to $\widehat R$, the accepted configuration, uncovered requirements, and the evidence memory. Every reorientation is followed by a correspondence check before the new configuration can authorize a scan, and every scan is subjected to deterministic acquisition, datum, association, fit, and admissibility checks before its result can support conformance. An admissible result is written to evidence memory. If the next uncovered requirement lies on a different surface, the manager issues \textsc{change-access} as a normal progression action and converts it into a VLA reorientation subgoal. A locally recoverable acquisition failure instead triggers \textsc{reacquire} in the current configuration. The loop terminates with a supported conformance verdict only when every requirement in $\widehat R$ has admissible evidence; exhaustion of the applicable execution or reacquisition budget produces an \textsc{unresolved} operational outcome and routes the part to manual review.

This organization aligns the runtime interfaces with the false-accept decomposition in Eq.~\eqref{eq:false-accept-bound}. Scope grounding is the component that can incur omission event $G$; accepted physical-to-specification associations can incur correspondence events $D_f$; and evidence acquisition, admission, and conformance evaluation can incur requirement-level events $A_r$. Measurement steering governs local reacquisition and coverage-driven progression, while the scope, correspondence, admissibility, coverage, and conformance gates determine whether the resulting evidence can support \textsc{pass}. Planning or manipulation failures exposed by runtime verification, or reflected in failed evidence admission or coverage, may increase execution cost or produce \textsc{unresolved}, but do not silently propagate into the conformance path.


\subsection{Specification-Constrained Task Manager}
\label{sec:method-planner}

The task manager converts the inspection instruction into an explicit evidence-acquisition contract. The structured specification in Eq.~\eqref{eq:specification} provides a finite vocabulary of feature, group, functional-role, surface-region, datum, requirement, and verdict-rule identifiers. A VLM first grounds the instruction to a set of task concepts $\widehat{\mathcal C}$, after which a deterministic specification expansion produces the leaf requirements that must be evidenced,
\begin{equation}
\widehat{\mathcal C}=g_{\theta}(I,S),
\qquad
\widehat R=\operatorname{Expand}_{S}(\widehat{\mathcal C})
\subseteq\mathcal R_{\mathrm{all}}.
\label{eq:method-grounding}
\end{equation}
Here, $g_{\theta}$ may select only identifiers explicitly declared in $S$; schema validation rejects malformed or unknown outputs. $\operatorname{Expand}_{S}$ then resolves a requested feature, group, or functional role into its declared leaf evidence scope, surface dependencies, cross-feature relations, and datum dependencies. Thus, a group name is a compact task instruction rather than an implicit source of new evidence: every resulting requirement and dependency is traceable to an explicit entry in $S$. The grounding VLM neither reads nor edits nominal values and tolerances; downstream deterministic acquisition and conformance modules consume them directly from $S$. An omitted valid task concept remains a scope error and is measured by $\gamma$.

The manager maintains the inspection state
\begin{equation}
z_t=(\widehat R,\widehat x_t,\widehat\mu_t,
\mathcal M_t,\mathcal U_t,d_t,\mathbf B_t^{\mathrm{rem}}),
\label{eq:manager-state}
\end{equation}
where $\widehat x_t\in\mathcal X_c\cup\{\bot\}$ is the accepted support-stable configuration, $\widehat\mu_t$ is the current surface correspondence table, $\mathcal M_t$ is the evidence memory, $\mathcal U_t$ is the uncovered requirement set, $d_t$ is the latest structured measurement diagnostic, and $\mathbf B_t^{\mathrm{rem}}$ stores the remaining execution and reacquisition budgets. Before the first scan, $d_t=\varnothing$; thereafter it stores the most recent outcome in Eq.~\eqref{eq:measurement-diagnostic}. A deterministic action router converts the inspection state into a typed next-action request,
\begin{equation}
g_t^{\mathrm{next}}
=\operatorname{Route}
\!\left(d_t,\mathcal U_t,\widehat x_t,
\widehat\mu_t,\mathbf B_t^{\mathrm{rem}},S\right).
\label{eq:action-routing}
\end{equation}
The request specifies an admissible action class, candidate target set, and access or acquisition constraints. It may represent normal coverage progression or diagnostic recovery. A VLM then reads a serialization of the state and next-action request and proposes a semantic action, while a deterministic dispatcher checks and instantiates it:
\begin{equation}
\begin{aligned}
\widetilde a_t={}&\pi_{\theta}
\!\left(\operatorname{serialize}(z_t,S,g_t^{\mathrm{next}})\right),\\
a_t={}&\operatorname{Dispatch}
(\widetilde a_t;z_t,S,g_t^{\mathrm{next}}).
\end{aligned}
\label{eq:semantic-dispatch}
\end{equation}
A \textsc{reorient} proposal names the task-specified regions that should become scannable and the access constraints that must be satisfied; a \textsc{scan} proposal names a verified region and the requirement identifiers assigned to it. For \textsc{reacquire}, the router deterministically restricts dispatch to a revised acquisition in the current configuration. For \textsc{change-access}, it restricts dispatch to a reorientation that exposes the next uncovered task region. The dispatcher accepts a proposal only when all identifiers exist in $S$, the requested action type agrees with $g_t^{\mathrm{next}}$, the required surface dependencies agree with $\widehat\mu_t$, and the action is legal in the runtime state machine. This is schema-validated semantic dispatch, not token-level constrained decoding: the VLM generates a structured proposal, which is parsed and rejected or retried if validation fails. 

Evidence coverage is audited deterministically after $\widehat R$ is fixed. For every $r\in\widehat R$, the manager queries $\mathcal M_t$ for admissible records associated with $r$ and every task-specified region on which it depends. Admitted records update $\mathcal U_t$, while rejected acquisitions update $d_t$; together they determine the next admissible action through Eq.~\eqref{eq:action-routing}. Detecting one nonconforming requirement does not stop acquisition of the remaining evidence. Natural-language interpretation is therefore isolated in $g_{\theta}$, while task expansion, action typing, action legality, evidence coverage, and termination remain reproducible operations over $S$ and $\mathcal M_t$.

\subsection{Active Surface Correspondence}
\label{sec:method-correspondence}

\begin{figure*}[t]
\centering
\includegraphics[width=\textwidth]{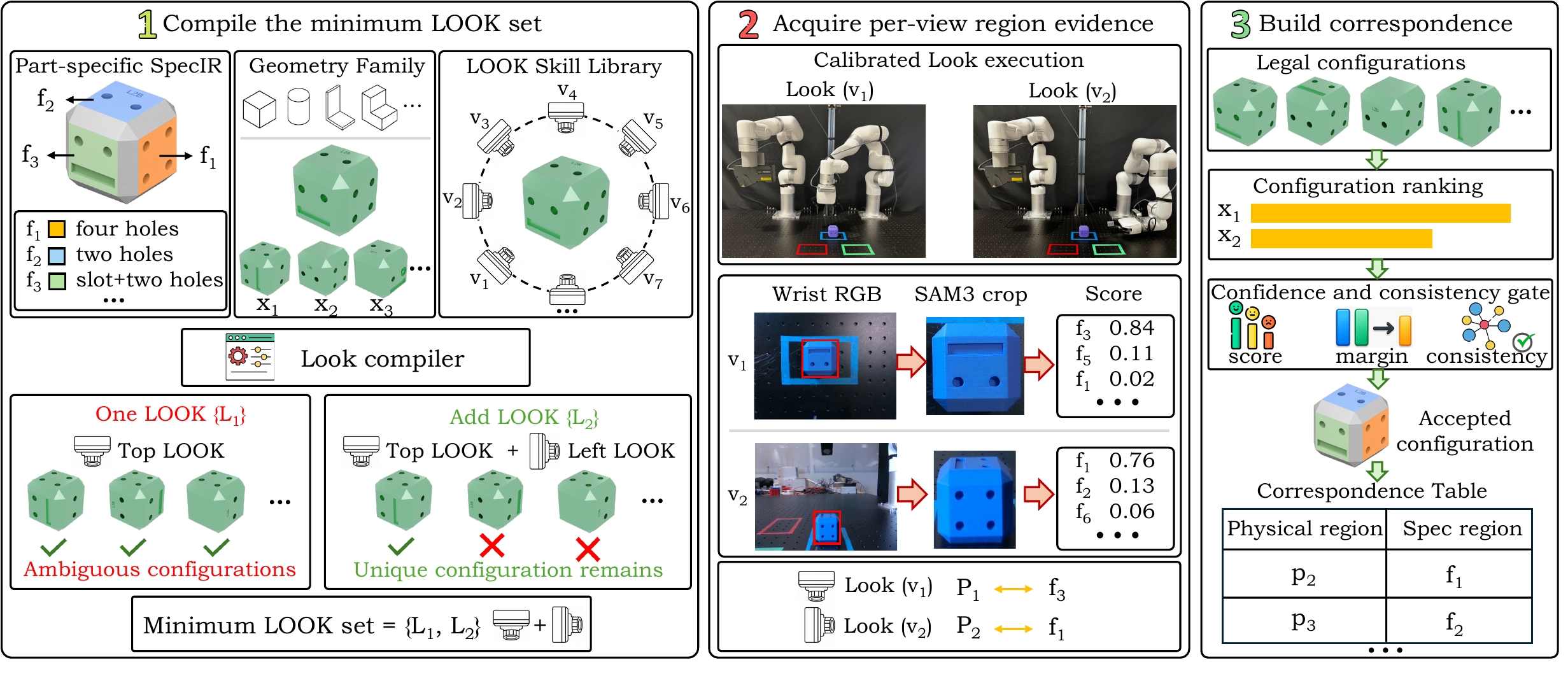}
\caption{Specification-driven construction of the surface-correspondence table. \emph{Step 1: Minimum LOOK-set compilation.} Given the part specification and its geometry-family LOOK library, the compiler selects the smallest subset that separates all non-equivalent stable configurations; in the cube example, one upper LOOK is ambiguous, whereas upper and left LOOKs identify one configuration. \emph{Step 2: Per-view region evidence acquisition.} The robot executes the selected calibrated LOOKs, SAM3 \cite{carion2026sam} crops the workpiece, and the VLM scores the candidate specification regions. \emph{Step 3: Correspondence construction.} The per-view scores rank the legal configurations. If the best candidate passes the score, top-two-margin, and anchor/topology gates, its physical-to-specification assignment is released as $\widehat\mu_t$; otherwise, no table is produced.}

\label{fig:surface-correspondence}
\end{figure*}

Surface correspondence bridges task-specified regions and the physical geometry from which evidence is acquired. A region may be a surface patch, an end region, an arm, a tread, a riser, or a member of a multi-region relation, while the part arrives in an unknown configuration from a finite support-stable set $\mathcal X_c$. A scan can therefore be numerically accurate yet support the wrong requirement if its physical source is assigned to the wrong region. Fig.~\ref{fig:surface-correspondence} summarizes the initial construction of this association. Each $x\in\mathcal X_c$ jointly encodes support contact, discrete orientation class, physically scannable regions, and a physical-to-specification region assignment.

The first stage determines which reusable LOOK skills are required before any image is acquired. For each supported geometry family $c$ and inspection zone $z$, we define a finite admissible library $\mathcal V_{c,z}^{\mathrm{adm}}=\{v_1,\ldots,v_K\}$. Each $v_i$ couples an inspection-zone-calibrated wrist-camera pose with a collision-free robot motion and a designated physical target region that is visually dominant in the acquired image. This library is calibrated once and reused across specification variants of the same geometry family.

For a particular part specification $S$, the compiler enumerates the finite support-stable configuration set $\mathcal X_c$. It constructs each specification-region appearance signature from the declared geometry and visible features, including feature type, count, and approximate layout. The family model separately supplies view-to-region visibility, adjacency and opposite relations, handedness, and specification-declared equivalence. Let $\Sigma_{\mathcal V}(x;S)$ be the joint region signature predicted when configuration $x$ is observed using LOOK subset $\mathcal V$. The compiler selects
\begin{equation}
\begin{aligned}
\mathcal V^{*}(S)
&=\underset{\mathcal V\subseteq\mathcal V_{c,z}^{\mathrm{adm}}}{\arg\min}\ |\mathcal V|\\
\mathrm{s.t.}\quad
&\Sigma_{\mathcal V}(x;S)\neq\Sigma_{\mathcal V}(x';S),
\quad \forall x\not\sim_S x'.
\end{aligned}
\label{eq:min-look-subset}
\end{equation}
where $x\sim_S x'$ denotes configurations that are inspection-equivalent under $S$. Thus, LOOK skills are calibrated at the geometry-family and inspection-zone level, but the minimum subset is compiled for the feature distribution of the particular part. Different specifications from the same family may therefore require different LOOK subsets. In the cube example in Fig.~\ref{fig:surface-correspondence}, Step~1, one upper LOOK is consistent with several yaw hypotheses; adding the left LOOK separates them and leaves one legal configuration.

As shown in Fig.~\ref{fig:surface-correspondence}, Step~2, for every selected LOOK skill $v_i\in\mathcal V^{*}(S)$, the manipulation expert executes the corresponding pre-calibrated motion and acquires wrist-camera image $o_i$. SAM3 grounds the workpiece and produces the image crop $c_i$, after which the VLM compares $c_i$ with the specification-derived description $d_S(f)$ of every region $f\in\mathcal F$:
\begin{equation}
\begin{aligned}
c_i&=\operatorname{Crop}\!\left(o_i,\operatorname{SAM3}(o_i,p_c)\right),\\
m_i(f)&=\operatorname{VLMMatch}\!\left(c_i,d_S(f)\right),
\end{aligned}
\label{eq:region-match-score}
\end{equation}
where $p_c$ is the object-family prompt and $m_i(f)\in[0,1]$ is a region-matching score. The VLM therefore provides per-view semantic evidence; it neither outputs a complete correspondence table nor selects among arbitrary poses.

As shown in Fig.~\ref{fig:surface-correspondence}, Step~3, the code constructs the table by enumerating every legal $x\in\mathcal X_c$. Let $F_i(x)\subseteq\mathcal F$ be the specification region or region set predicted at observation slot $i$ under $x$. Each configuration receives the normalized weighted score
\begin{equation}
s_t(x)=
\frac{\displaystyle\sum_{v_i\in\mathcal V_t}
w_i\max_{f\in F_i(x)}m_i(f)}
{\displaystyle\sum_{v_i\in\mathcal V_t}w_i},
\label{eq:configuration-score}
\end{equation}
where $\mathcal V_t\subseteq\mathcal V_{c,z}^{\mathrm{adm}}$ is the set of executed LOOK skills and $w_i$ is a fixed reliability weight for skill $v_i$. Let $x_t^{(1)}$ and $x_t^{(2)}$ be the highest- and second-highest-scoring non-equivalent configurations. The system accepts $\widehat x_t=x_t^{(1)}$ only if
\begin{equation}
\begin{aligned}
s_t\!\left(x_t^{(1)}\right)&\geq\eta_s,\\
s_t\!\left(x_t^{(1)}\right)-s_t\!\left(x_t^{(2)}\right)&\geq\eta_m,\\
A_t\!\left(x_t^{(1)};S\right)&=1,
\end{aligned}
\label{eq:correspondence-gate}
\end{equation}
where $A_t$ is a deterministic anchor and topology-consistency check. In the cube implementation, for example, it rejects a configuration that conflicts with a sufficiently strong upper-region anchor; the finite configuration model already enforces opposite, adjacency, handedness, and exclusivity constraints. The specification-compiled LOOK set is executed once for the correspondence decision. If any gate fails, $\widehat x_t=\bot$, no correspondence table is released, and the runtime abstains rather than acquiring an additional LOOK or guessing.

Every accepted configuration defines a complete physical-to-specification assignment $M_x$. The runtime table is therefore
\begin{equation}
\widehat\mu_t(p)=M_{\widehat x_t}(p),
\qquad p\in\mathcal F^{\mathrm{phys}},
\label{eq:estimated-correspondence}
\end{equation}
with observed entries marked as image-supported and unobserved entries marked as configuration-derived. The latter are completed only through the accepted finite configuration and relations explicitly encoded by $S$. Globally symmetric configurations need only be resolved up to specification-declared equivalence when all members induce the same required evidence. Every admitted measurement record stores the physical source region, associated specification region, configuration identifier, observation sources, and gate scores in traceability field $\ell$. 


\subsection{Manipulation Expert}
\label{sec:method-manipulation}

\begin{figure}[t]
\centering
\includegraphics[width=\columnwidth]{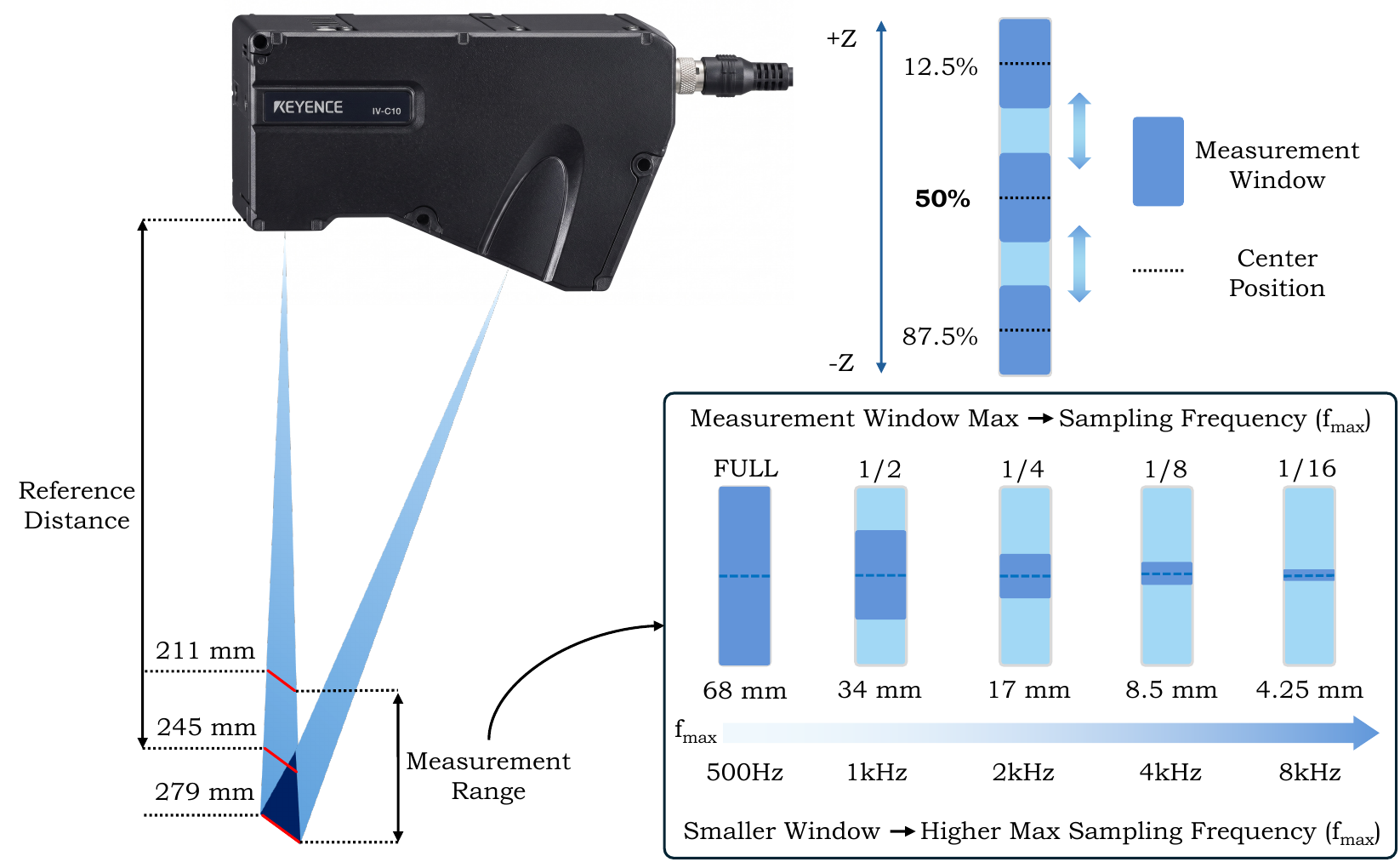}
\caption{Measurement-range configuration of the Keyence LJ-X8200 laser profiler. The measurement window can be shifted along the $Z$-axis by adjusting its center position and reduced from the full 68~mm range to 34, 17, 8.5, or 4.25~mm. Reducing the measurement range increases the maximum sampling frequency $f_{\max}$ from 500~Hz to 8~kHz.}
\label{fig:scanner}
\end{figure}

The manipulation expert exposes three atomic skills: \textsc{look}, \textsc{reorient}, and \textsc{sort}. \textsc{Look} executes a pre-calibrated camera action $v\in\mathcal V^{*}(S)$ and returns an observation for surface-correspondence estimation. The other two skills are implemented by a VLA policy. \textsc{Reorient} changes physical access through geometry-dependent grasp, flip or roll, regrasp, and placement actions. \textsc{Sort} performs the final pick-and-place into the pass or no-pass region, but is enabled only after the conformance evaluator authorizes a verdict. This interface lets the task manager select the semantic action and target while leaving contact-rich execution to the learned policy.

When an uncovered requirement lies outside the currently scannable region set, the task manager issues \textsc{change-access} with the target specification region and its access constraints. The manipulation expert serializes this request as a language-conditioned VLA subgoal and invokes \textsc{reorient}; the target is determined by the uncovered evidence scope rather than by a generic manipulation objective. We use real-time chunking (RTC) for continuous VLA execution~\cite{black2026real}, with a calibrated per-skill time limit after which control returns to the task manager~\cite{hu2026matters}. These are execution details rather than success criteria: the outcome of a VLA invocation is accepted only through the independent checks below.

The VLA-based \textsc{reorient} skill is treated as an unverified physical transition until its outcome is observed. After each invocation, the task manager applies the commanded reorientation to the previously accepted configuration and correspondence, yielding a candidate post-reorientation state. The manipulation expert then executes the specification-compiled \textsc{look} skills. The candidate is committed as the updated configuration $\widehat x_{t+1}$ and correspondence table $\widehat\mu_{t+1}$ only if these observations pass the surface-correspondence gate and verify that the requested region $f$ belongs to the resulting set of scannable specification regions. Only then is a scan of $f$ authorized. Thus, independent verification is realized by composing the two skills as \textsc{reorient}$\rightarrow$\textsc{look}, rather than by trusting the VLA-predicted outcome.

If the observed scannable region is not the requested target or the compiled LOOK set otherwise fails the correspondence gate, the candidate update is not committed and no scan is authorized. The manager may issue another legal \textsc{reorient} attempt within the execution budget; otherwise the episode becomes \textsc{unresolved} and the part is routed to manual review. A timeout, dropped part, invalid placement, or unintended reorientation is likewise handled as an unverified transition and cannot directly produce measurement evidence.

This interface separates correctness from action efficiency. A poor plan, failed grasp, or unintended flip increases the number of observations and reorientation attempts, but affects false-accept risk only if an incorrect correspondence also passes the mandatory post-action gate, which is counted by $\delta$. The VLA may therefore flexibly realize \textsc{reorient}, while the calibrated \textsc{look} skill and correspondence gate independently determine whether the resulting configuration may support measurement evidence.

\subsection{Measurement Expert}
\label{sec:method-measurement}

\begin{figure}[t]
\centering
\includegraphics[width=\columnwidth]{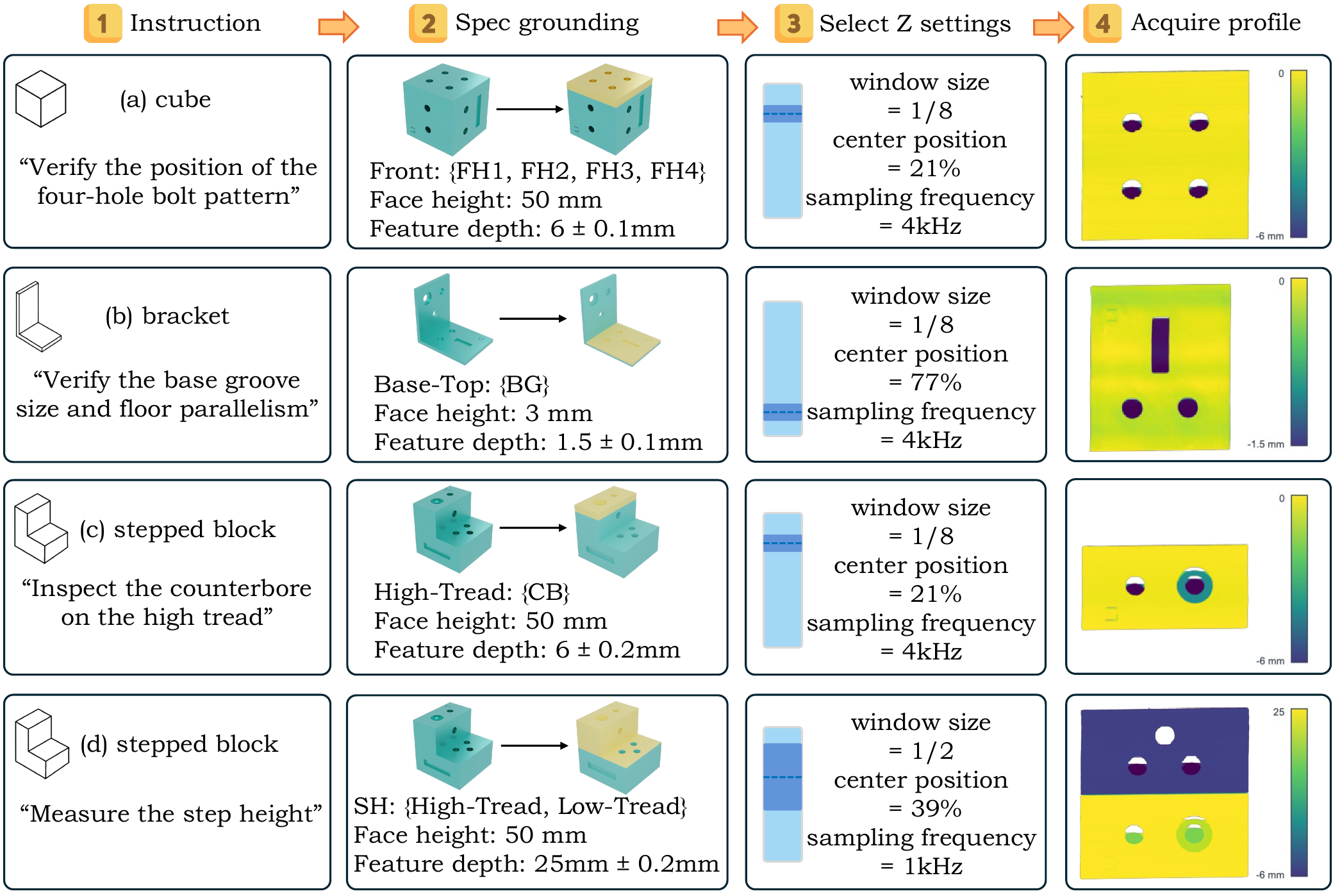}
\caption{Examples of specification-driven vertical measurement-window selection for different objects and inspection requests. The instruction is grounded to the required surface, feature depth or height, and tolerance before the expert selects the measurement $Z$ range and center position. The displayed sampling frequency is the maximum frequency $f_{\max}(R_f)$ supported by the selected $Z$ range, rather than the commanded scan frequency. (a) The cube bolt pattern uses a narrow range centered near the raised surface. (b) The shallow groove on the bracket uses the same range size but a shifted center because its base surface is at a different height. (c) The counterbore on the high tread again uses a narrow, surface-centered range. (d) Step-height inspection must contain both height levels, requiring a wider range, an intermediate center, and consequently a lower maximum sampling frequency. The acquired profiles confirm that the task-required geometry lies within the configured window.}
\label{fig:z-example}
\end{figure}

\begin{figure*}[t]
\centering
\includegraphics[width=\textwidth]{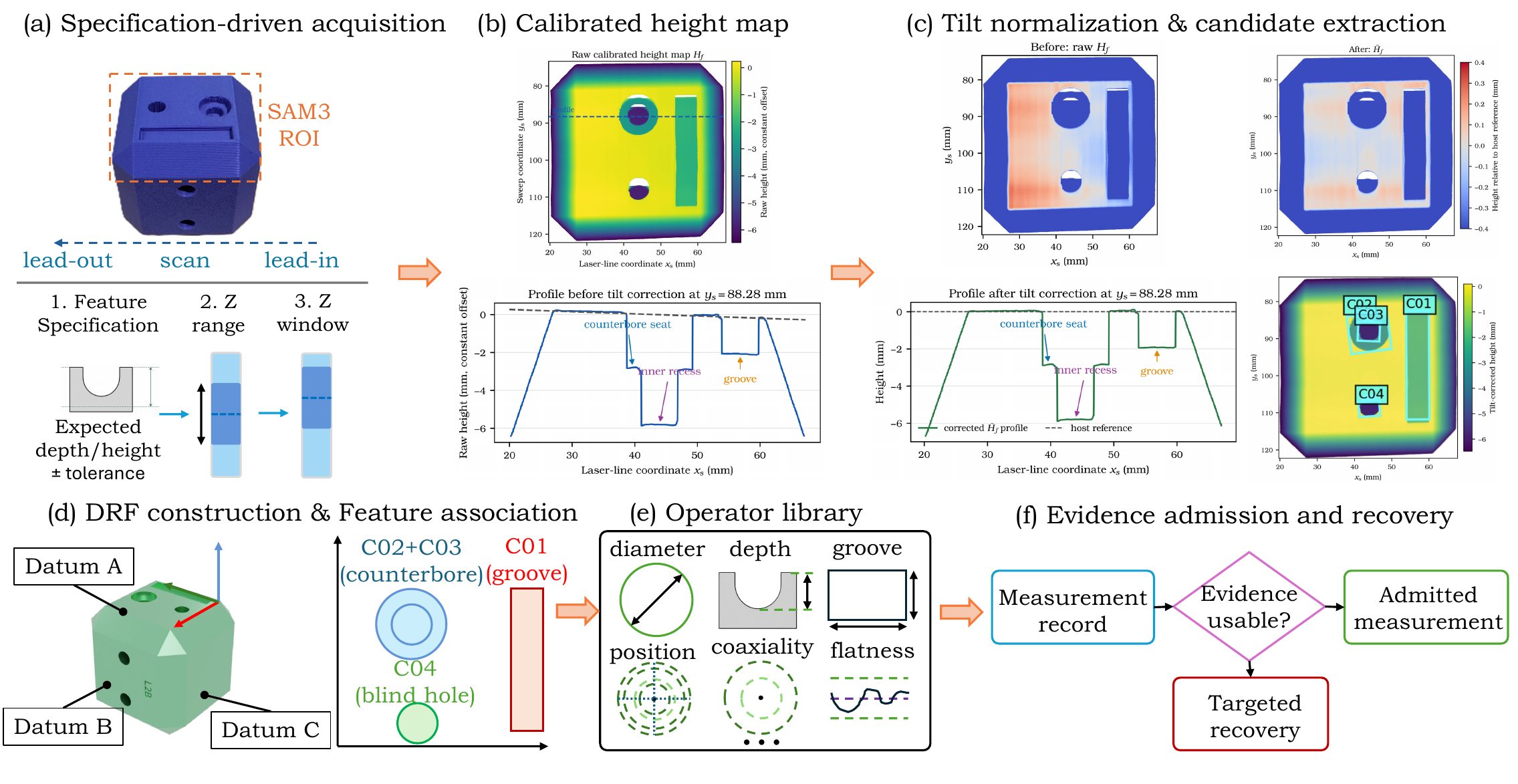}
\caption{Measurement Expert pipeline from a verified inspection request to admissible evidence. (a) The grounded specification defines the feature region, lateral scan path, and vertical measurement window. (b) Calibrated laser profiles are registered into a height map. (c) Host-surface tilt is normalized before geometric candidates are extracted. (d) The datum reference frame is constructed and candidates are deterministically associated with the requested specification features. (e) A requirement-specific operator computes the reported geometric quantity. (f) Deterministic admissibility predicates either admit the measurement to Evidence Memory or trigger targeted reacquisition; an inadmissible record cannot support conformance.}
\label{fig:measurement-expert}
\end{figure*}

The measurement expert maps a verified scannable-region request $(f,\mathcal R_f)$ to traceable numerical evidence. Here, $f$ is a task-specified region and $\mathcal R_f\subseteq\widehat R$ is the set of requirements assigned to it. Multi-region requirements are decomposed into linked acquisitions whose partial evidence is retained until all referenced regions are covered. For each $r\in\mathcal R_f$, a requirement-type-specific operator produces
\begin{equation}
\widehat q_r=
h_{\operatorname{type}(r)}
\!\left(H_f,\mathrm{DRF}_r,j^{*}(r)\right),
\label{eq:measurement-operator}
\end{equation}
where $H_f$ is the registered height data, $\mathrm{DRF}_r$ is the datum reference frame required by $r$, and $j^{*}(r)$ is the deterministically associated feature or supporting surface. As summarized in Fig.~\ref{fig:measurement-expert}, the expert first configures and acquires a calibrated scan, reconstructs geometry in the declared datum frame, associates candidates without using their conformance values, and applies a deterministic operator and admissibility predicate. It returns the requirement identity, measured quantity, admissibility status, and traceability metadata rather than a free-form interpretation or verdict.

\subsubsection{Specification-Driven Acquisition}

After the correspondence gate authorizes region $f$, SAM3 localizes its lateral acquisition window in the inspection-camera image. The highest-confidence mask satisfying calibrated safe-region constraints is selected. The mask extent, enlarged by a fixed margin, is mapped through a calibrated planar homography to the measurement-arm plane and projected onto the profiler's calibrated sweep axis. Lead-in and lead-out segments ensure that the requested interval is traversed at constant speed, and the entire path must remain inside the calibrated workspace and inspection safe region. 

The lateral path is coupled with a specification-driven vertical window. Let $D_s=D_T-h_o$ be the exposed support-surface distance, obtained from the calibrated scanner-to-table distance $D_T$ and the specified or observed object height $h_o$. The requirement defines an allowable feature envelope $[D_f^-,D_f^+]$, enlarged by its tolerance and acquisition margin $m_z$ so that a nonconforming feature is not clipped. From the profiler's discrete ranges $\{R_k\}$, the expert selects
\begin{equation}
\begin{aligned}
R_f&=\min\left\{R_k:
R_k\geq D_f^+-D_f^-+2m_z\right\},\\
D_f^c&=(D_f^-+D_f^+)/2,
\end{aligned}
\label{eq:vertical-range}
\end{equation}
and converts $D_f^c$ to the device center command after checking the device's center limits. Fig.~\ref{fig:z-example} shows that this rule shifts the same narrow range across surfaces of different heights and selects a wider range when both levels of a step must be observed.

The selected range constrains the maximum profile frequency $f_{\max}(R_f)$. Given the commanded sweep pitch $p_v$, the expert chooses $f_s\leq f_{\max}(R_f)$ and sets
\begin{equation}
v_s=f_s p_v,
\qquad
f_s\leq f_{\max}(R_f),
\label{eq:scan-pitch}
\end{equation}
subject also to the robot-speed limit. The transverse profiler pitch and $p_v$ define the sampling grid and are validated before motion; an infeasible configuration is rejected rather than treated as a low-quality measurement afterward. Streamed profiles are stacked into a registered height map $H_f\in(\mathbb R\cup\{\varnothing\})^{N_v\times N_u}$, where $\varnothing$ denotes a missing or out-of-range return. Each map retains the path, profiler pose, calibration identifier, acquisition windows and settings, timestamp, and requested region.

\subsubsection{Calibrated Geometric Reconstruction}

Residual support-surface tilt is removed before height-level decomposition and candidate extraction. A robust RANSAC plane is fitted to spatially distributed support samples and subtracted from valid cells of $H_f$; missing returns remain undefined. The normalized map is used to separate height populations, while the fitted plane, inlier support, and residual statistics are stored with the scan. Insufficient support or excessive residual error makes the dependent evidence inadmissible and requests a larger verified support patch.
Measurements are then expressed in the datum reference frame (DRF) declared by each requirement rather than aligned freely to nominal geometry. Given the ordered datum definition $\Delta_r$, a deterministic constructor fits its primitives in precedence order and maps calibrated samples $\mathbf p_i$ into datum coordinates:
\begin{equation}
\mathrm{DRF}_r=\Phi_{\Delta_r}(H_f,S),
\qquad
(u_i,v_i,w_i)^{T}=\mathrm{DRF}_r(\mathbf p_i).
\label{eq:drf-construction}
\end{equation}
A planar primary datum supplies the origin and normal, a projected secondary direction fixes the in-plane axes, and a tertiary datum fixes the remaining translation; edge and axis datums use their fitted directions with the declared origin and clocking constraints. The expert validates point count and spread, fit residuals, constrained-degree observability, axis consistency, and datum availability before measuring a feature. A failed datum check makes all dependent measurements inadmissible, preventing a plausible fit in an invalid frame from supporting a GD\&T result.

\subsubsection{Deterministic Association and Operator Library}

Feature association uses task-specified geometry and topology in the DRF, not the value of the characteristic being inspected. Registered scan patches are segmented into candidate supporting surfaces, openings, annular recesses, grooves, steps, and residual recessed components using height discontinuities, connectivity, and their relation to fitted support regions. For an opening, the boundary at which the supporting surface loses valid support defines the \emph{opening contour}; fitting this contour avoids bias from partially observed walls or bottoms.

Let $\boldsymbol\xi_j$ denote the DRF descriptor of candidate $j$---for example, a point, axis, contour anchor, or surface-patch coordinate---and let $\boldsymbol\xi_r^0$ be the basic descriptor of the specification target referenced by requirement $r$. The specification also declares the required candidate primitive type $\operatorname{ftype}(r)$. The target candidate is selected by
\begin{equation}
\begin{aligned}
j^{*}(r)
&=\underset{j\in\mathcal J_f}{\arg\min}\ 
d_r(\boldsymbol\xi_j,\boldsymbol\xi_r^0)\\
&\mathrm{s.t.}\quad
\operatorname{type}(j)=\operatorname{ftype}(r),\\
&\hphantom{\mathrm{s.t.}}\quad
\boldsymbol\xi_j\in\mathcal N_r,
\end{aligned}
\label{eq:feature-association}
\end{equation}
where $\mathcal J_f$ is the set of geometrically valid candidates on region $f$, $\mathcal N_r$ is the association neighborhood declared by $S$, and $d_r$ is a primitive-appropriate spatial distance defined independently of the characteristic under test. For a circular opening on a planar patch, this reduces to the nearest basic center in $(u,v)$; axis-based relations compare fitted axes. Because lookup does not use measured depth, diameter, form error, or conformance status, an out-of-tolerance feature is not replaced by a different candidate merely because the latter is numerically closer to nominal.

The operator library evaluates the associated geometry using fixed robust losses and quantile levels. Let $C_f$ be an opening contour, $\Omega_f$ its feature samples, and $\pi_A$ the host plane fitted after excluding designed features, boundaries, and missing returns. With radial residual $\eta_i(u_c,v_c,r)=\sqrt{(u_i-u_c)^2+(v_i-v_c)^2}-r$, a circular opening is measured by
\begin{equation}
\begin{aligned}
(\widehat u_c,\widehat v_c,\widehat r)
&=\underset{u_c,v_c,r}{\arg\min}
\sum_{i\in C_f}\rho\!\left(\eta_i(u_c,v_c,r)\right),\\
\widehat q^{\mathrm{diam}}&=2\widehat r.
\end{aligned}
\label{eq:opening-diameter}
\end{equation}
and a recessed-feature depth is the median host-to-feature separation
\begin{equation}
\widehat q^{\mathrm{depth}}
=Q_{0.50}\!\left(
\{\pi_A(u_i,v_i)-w_i:(u_i,v_i,w_i)\in\Omega_f\}
\right).
\label{eq:feature-depth}
\end{equation}
A through opening reports no bottom depth. Counterbores apply the same two operators to nested contours and observed levels, yielding separate outer/inner diameters and seat/inner depths. Grooves use quantile extents along the major and minor footprint axes for length and width and Eq.~\eqref{eq:feature-depth} for depth; step height is the absolute median offset between two support regions in a common DRF.

Form, orientation, and location operators reuse the same validated geometry. Projected true position is twice the DRF-plane distance between the fitted and basic centers, while projected counterbore coaxiality is twice the distance between the fitted outer and inner centers. Flatness is the robust peak-to-valley band of host-plane residuals, and groove-floor parallelism is the corresponding height band relative to the host plane. These are explicitly defined operational measures rather than claims to compute ISO minimum-zone quantities.
Cross-feature consistency and spacing are respectively the absolute difference between like-for-like admitted values and the Euclidean distance between admitted DRF centers. Recessed-defect depth is the largest robust positive-residual depth after designed features, missing returns, and an edge band are excluded. All operators act on calibrated points in a validated DRF, and relation operators run only after their constituent records are admitted.

\subsection{Evidence-Coverage Steering and Diagnostic Recovery}
\label{sec:method-recovery}

A numerical value is admitted only when the acquisition and fitted geometry support the requested characteristic. Common gates verify sufficient valid-return density, calibration and safe motion, absence of measurement-range clipping, valid tilt compensation, valid DRF construction, a unique candidate in $\mathcal N_r$, and bounded primitive-fit residuals. Characteristic-specific gates then test the necessary observable geometry: opening diameter requires adequate arc coverage, radial inliers, contour closedness, and no image-boundary contact; depth requires valid host and feature samples separated into stable height populations; a counterbore requires two ordered contours and the requested observed level; groove dimensions require a nontruncated connected footprint and a valid floor fit; step height requires both support regions; projected true position requires a valid fitted center and every referenced datum; flatness requires sufficient interior coverage after edge and feature exclusion; projected coaxiality requires valid nested-circle fits; parallelism requires sufficient floor extent; and relation operators require every constituent measurement to be admitted. For each requirement type, these checks form a deterministic predicate $\mathcal A_r$, and the expert sets
\begin{equation}
b_r=\mathbb{I}\!\left[
\mathcal A_r(H_f,\mathrm{DRF}_r,j^{*}(r),\widehat q_r,r)=1
\right]
\label{eq:admissibility-test}
\end{equation}
and never uses a record with $b_r=0$ to support \textsc{pass}. Each predicate corresponds to an observable physical or numerical failure that can be reproduced from the stored scan; it is not a generic model-confidence score. A record with $b_r=1$ produces diagnostic \textsc{admitted}; otherwise, the failed predicates and remaining recovery budget determine either \textsc{reacquire} or \textsc{unresolved}.

Panel (f) of Fig.~\ref{fig:measurement-expert} summarizes targeted acquisition recovery at a high level. \textsc{Reacquire} retains the accepted configuration and adjusts only the measurement process: boundary contact expands the SAM3-derived lateral patch, vertical clipping shifts the device center command corresponding
to $D_f^c$ or selects a larger $R_f$, excessive missing returns revise the exposure or measurement window, and a locally unstable tilt or primitive fit requests a larger support patch or repeated acquisition. The revised scan is processed by the same measurement operator and must pass the same predicate $\mathcal A_r$ before it can enter evidence memory.
After evidence for the current surface is admitted, the manager recomputes $\mathcal U_t$. If requirements remain on another surface, the action router issues \textsc{change-access} as the next planned action, and the dispatcher instantiates the corresponding structured VLA subgoal. The resulting configuration must pass the correspondence and scan-authorization gates before the next surface is measured. If local reacquisition exhausts its configured budget, the diagnostic becomes \textsc{unresolved}. Every attempt, including rejected attempts, is appended to the traceability log, while only \textsc{admitted} records contribute to evidence coverage.

Recovery is therefore evidence-driven rather than value-driven. Because the sampling tuple is validated before motion, insufficient nominal sampling density is not a normal post-acquisition recovery case. Likewise, an admitted measurement outside the specification limit is retained as measured nonconformance and does not trigger reacquisition or change the planned surface progression. Reacquisition is requested only for inadmissible evidence; after the configured recovery budget is exhausted, the system records incomplete evidence, returns \textsc{unresolved}, routes the part to manual review, and cannot release it as \textsc{pass}.

\subsection{Specification-Driven Conformance Evaluation}
\label{sec:method-evaluation}

Conformance evaluation is deliberately separated from task planning so that a VLM proposal cannot turn incomplete or ambiguous evidence into a \textsc{pass} verdict. After the evidence-admissibility predicates have passed, a bilateral dimensional requirement with nominal value $q_r^0$ and half-width tolerance $\tau_r$ is evaluated by
\begin{equation}
|\widehat q_r-q_r^0|\leq\tau_r,
\label{eq:direct-bilateral}
\end{equation}
and a nonnegative GD\&T characteristic with upper limit $\tau_r$ is evaluated by
\begin{equation}
\widehat q_r\leq\tau_r.
\label{eq:direct-upper}
\end{equation}
Evidence outside the task-specified limit is marked nonconforming. Missing coverage or inadmissible evidence is processed by the action router and cannot support a \textsc{pass} decision. This direct comparison allows a new nominal value or tolerance to be supplied through the structured specification without training a classifier or authoring a part-specific inspection program.
Each admitted measurement is appended to $\mathcal M_t$ together with its acquisition-time correspondence snapshot $\widehat\mu_e$, scan identifier, and fit diagnostics; the evaluator then attaches its deterministic requirement status. Let $e\models_{\widehat\mu_e}r$ denote that the record contains the acquisition primitives required by $r$ and that their observed physical regions are associated by $\widehat\mu_e$ with every specification region on which $r$ depends. Runtime coverage is complete only if every requirement in $\widehat R$ has such an admitted record,
\begin{equation}
\begin{aligned}
\widehat{\operatorname{Cov}}(\mathcal M_t,\widehat R)=1
\ \Longleftrightarrow\ 
&\forall r\in\widehat R,\ \exists e\in\mathcal M_t:\\
&b(e)=1\ \wedge\ \widehat r(e)=r
\ \wedge\ e\models_{\widehat\mu_e}r.
\end{aligned}
\label{eq:runtime-coverage}
\end{equation}
This runtime test uses the estimated correspondence snapshot stored with each record and is therefore distinct from the latent ground-truth coverage criterion used for evaluation. Missing evidence causes continued acquisition or, after recovery is exhausted, an unresolved outcome; it is never interpreted as evidence of conformance.

After coverage is complete, the evaluator applies the specification rule $\Psi$ to the feature statuses and group relations to produce $\widehat y_{\mathrm{conf}}$ and a complete inspection report; the manager then sets the supported operational outcome $\widehat y$. The aggregation is a deterministic program over requirement identifiers and cannot alter the underlying measurements. Only then does the manager authorize the manipulation expert to sort the part. Hence a released \textsc{pass} must have traversed three explicit gates: the task requirement must be present in $\widehat R$, its evidence must pass surface-correspondence and admissibility checks, and its deterministic status must satisfy the specification.

\section{Experimental Setup} 

\subsection{Hardware Platform}

\begin{figure}[t]
\centering
\includegraphics[width=0.8\columnwidth]{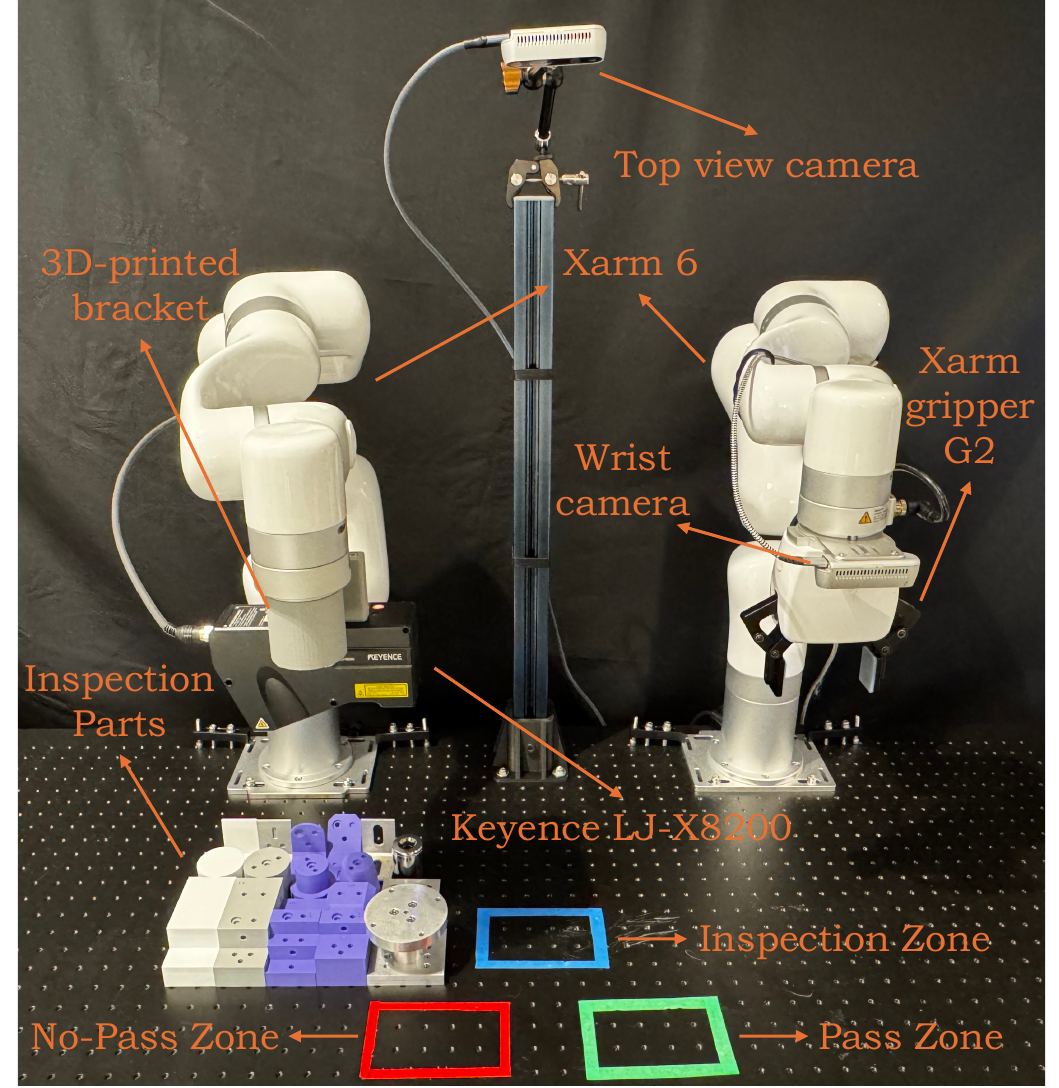}
\caption{Our real-world experiment setup. One XArm 6 carries a Keyence LJ-X8200 laser line profiler, while the other is equipped with a G2 gripper and wrist camera for part manipulation and local observation. A fixed top-view camera provides global workspace observations. Parts are measured in the inspection zone and subsequently sorted into the pass or no-pass zone.}
\label{fig:setup}
\end{figure}

The platform (Fig.~\ref{fig:setup}) consists of two six-degree-of-freedom UFACTORY xArm~6 manipulators on a shared optical breadboard, with overlapping workspaces over a central inspection region. Their heterogeneous end-effectors realize the dual heterogeneous experts design in hardware.
The \emph{measurement arm} carries a Keyence LJ-X8200 laser line profiler on a 3D-printed bracket, driven by a Keyence LJ-X8000A controller. The profiler delivers $1~\mu\mathrm{m}$ Z-axis repeatability and a $25~\mu\mathrm{m}$ X-axis profile interval over 3200 points per profile at a $245~\mathrm{mm}$ standoff. This arm is hand-eye calibrated and its tooling is never changed, preserving the calibration throughout each inspection cycle.
The \emph{manipulation arm} carries a UFACTORY xArm Gripper~G2 parallel-jaw gripper and a wrist-mounted Intel RealSense D435i, and performs all pick, place, flip, and reorientation actions. A second D435i on a fixed central mast provides the global top-view observation. The manipulation policy observes both the wrist and top-view cameras; the planner and the object-grounding stage use the top-view image alone.
The breadboard defines three regions for the tolerance-conditioned sorting task: an \emph{inspection zone} where the workpiece is measured, and a \emph{pass zone} and \emph{no-pass zone} into which the manipulation arm sorts it by the final judgment.

\subsection{Implementation Details}
\label{sec:implementation-details}

We instantiate the VLM with
Qwen3-VL-8B-Instruct \cite{Qwen3-VL}, which is used for instruction
grounding, semantic action proposal, and visual interpretation in the
surface-correspondence module. The manipulation expert is instantiated
with the $\pi_{0.5}$\cite{intelligence2025pi05visionlanguageactionmodelopenworld}.
A single shared $\pi_{0.5}$ policy supports all workpiece geometries
and manipulation skills, with each skill invoked using a distinct
language prompt.
Robot demonstrations are collected through a Meta Quest~3
teleoperation interface. For each workpiece geometry, we define the
reorientation skills required to transition among its supported stable
configurations. For example, the block requires backward-flip and
rightward-flip skills, whereas the bracket requires a flip-over skill.
We collect 60 teleoperated episodes for every geometry--skill pair,
including each geometry-specific reorientation skill and the final
pick-and-place skill. The demonstrations from all geometries and skills
are pooled to adapt the shared $\pi_{0.5}$ policy using low-rank
adaptation (LoRA) \cite{hu2021loralowrankadaptationlarge}. Fine-tuning is performed for 100,000 optimization
steps on a single NVIDIA RTX PRO~6000 Blackwell Workstation Edition
GPU. Intermediate checkpoints are evaluated in separate real-robot
validation trials, and the best-performing checkpoint is frozen before
benchmark evaluation.

\subsection{Benchmark Design} \label{sec: benchmark}

\begin{table*}[t]
\centering
\caption{Inspection task taxonomy and its industrial motivation. Manipulation difficulty is evaluated independently through reorientation cost and regret.}
\label{tab:task-taxonomy}
\setlength{\tabcolsep}{4pt}
\renewcommand{\arraystretch}{1.15}
\begin{tabular}{l p{0.15\textwidth} p{0.36\textwidth} p{0.37\textwidth}}
\toprule
\textbf{Task} & \textbf{Category} & \textbf{Industrial need} & \textbf{Required evidence} \\
\midrule
$T_1$ & Specified feature & Verify a named feature after machining or printing & Measure required characteristic of the requested feature \\
$T_2$ & Anomaly detection & Protect mating, sealing, bearing, or datum surfaces & Measure damage and form error on the requested surface \\
$T_3$ & Relation or GD\&T & Ensure alignment, interchangeability, and compatibility & Evaluate relations among features, surfaces, and datums \\
$T_4$ & Composite inspection & Inspect a declared feature group or interface & Aggregate all requirements in the declared group \\
$T_5$ & Functional goal & Verify an intent-level production goal
& Infer the required scope and collect complete evidence \\
\bottomrule
\end{tabular}
\end{table*}

The benchmark unit is an inspection episode comprising a workpiece design and specimen, an independently measured reference vector, a structured specification, a natural-language instruction, and an initial stable pose. This unit supports paired evaluation from identical initial conditions without treating repeated scans of one specimen as independent manufactured parts.

The benchmark uses four base geometries---cube, cylinder, L-bracket, and stepped block---and the five task families in Table~\ref{tab:task-taxonomy}. $T_1$ requests a named feature, $T_2$ a surface anomaly or form check, and $T_3$ a within- or cross-surface relation or GD\&T constraint. $T_4$ aggregates the requirements of a declared interface or part, whereas $T_5$ states a functional goal whose complete requirement scope must be inferred from the specification.
The object levels represent manufacturing progression rather than task difficulty. L0 contains four feature-free bodies used only to train grasping, reorientation, and placement. L1 adds holes, counterbores, grooves, datums, and functional groups and forms the main inspection benchmark. L2A and L2B further modify global morphology while retaining applicable L1 inspection roles and semantics; all level-specific feature, datum, position, and topology changes are declared in the corresponding specification. Thus, L2 evaluates transfer within a manufacturing lineage, not to unrelated categories. The real-object study is separate from this hierarchy.
Each design has one 3D-printed specimen whose as-built geometry is independently characterized with dedicated reference instruments. The resulting $m^{\mathrm{ref}}$ contains every inspectable characteristic; robot estimates are compared with these values rather than CAD nominals, and reference verdicts are obtained by applying the structured specification to $m^{\mathrm{ref}}$.
Each task family provides two instructions per object with different evidence scopes. All comparisons use frozen episode manifests; repeated executions are treated as repeated measurements, and decision rates are reported over task--specification pairs rather than independent manufactured specimens.

\begin{figure}[t]
\centering
\includegraphics[width=\columnwidth]{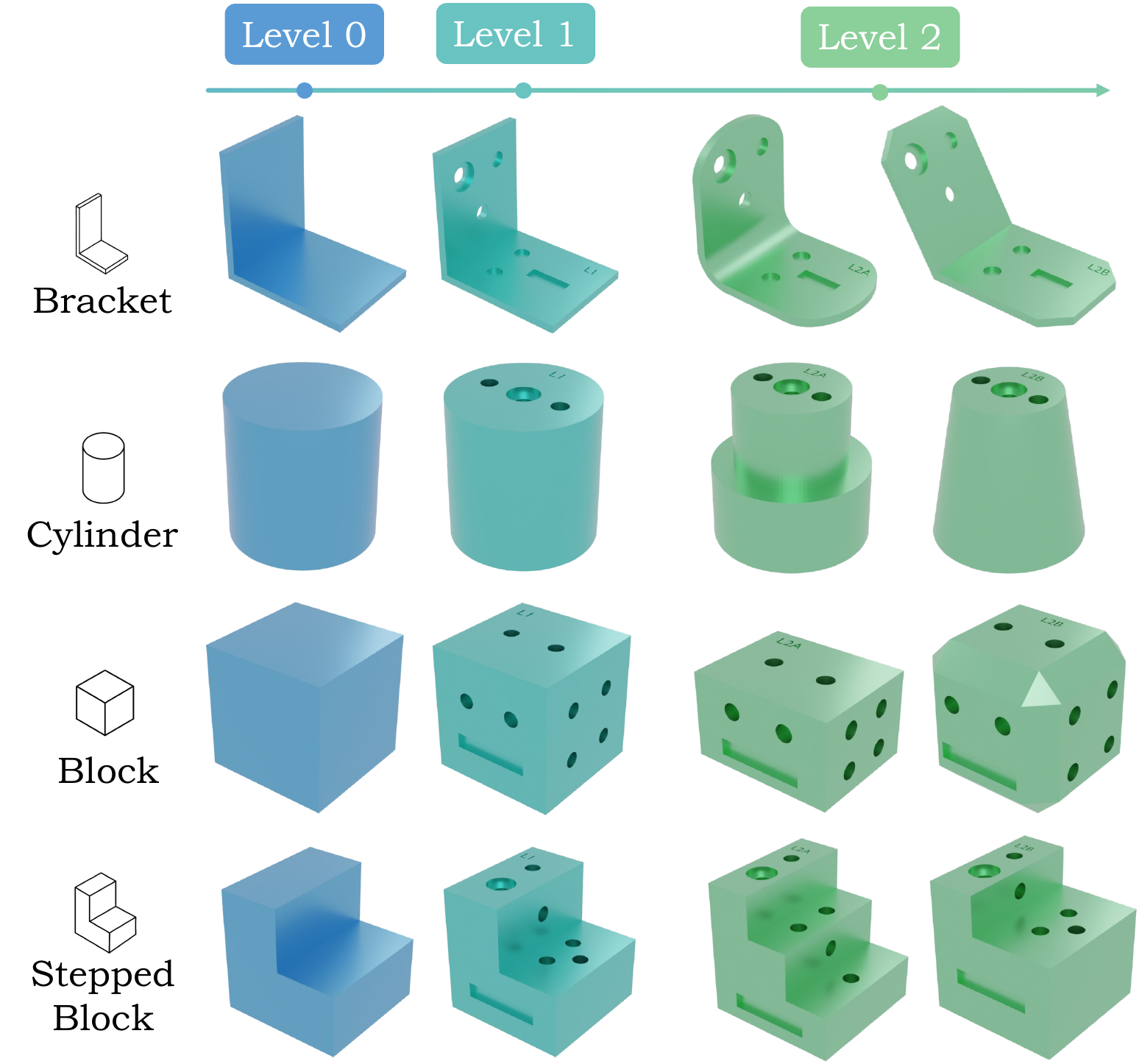}
\caption{Designed 3D-printed benchmark objects across increasing geometric variation. Four object families, bracket, cylinder, block, and stepped block, are organized into three levels. Level 0 provides the nominal geometries used for manipulation policy training, Level 1 introduces feature-level variation while largely preserving the overall shape, and Level 2 introduces larger morphology changes, with two variants per family.}
\label{object}
\end{figure}

\subsection{Compared Methods}
\label{sec:experimental-baselines}

We compare FRAME with three alternatives under matched parts, specifications, instructions, sensors, cameras, and episode manifests. The autonomous methods additionally share the robot action interface, frozen VLA manipulation policy, post-reorientation observations, laser acquisition interface, low-level geometric operators, and execution and recovery budgets; none receives oracle pose, surface correspondence, or measurement values, and failure to return a valid binary verdict is recorded as unresolved. \emph{Part and Task Specific Inspection} is a conventional operator-driven workflow in which a human interprets each part--task request, positions the part, configures Keyence regions, tools, and limits, executes the scans, and prepares the report; because it is not autonomous, it serves only as an engineered task-completion-time reference. \emph{Full Part Exhaustive Inspection} uses the shared VLA and automated metrology, admissibility tests, and deterministic comparisons, but follows a reusable per-part routine that exposes every registered surface and scans every registered feature before extracting the requested report, thereby isolating fixed exhaustive acquisition from task-specified acquisition. \emph{VLM without FRAME} uses one VLM to select observation, reorientation, scanning, and measurement actions and to determine completion, report, and verdict while retaining the shared VLA, calibrated observations, laser interface, and geometric fitting operators. It removes FRAME's deterministic specification expansion, authoritative surface correspondence, requirement-specific evidence admission, evidence memory, coverage audit, and specification-driven conformance evaluator, thereby testing whether VLM-level reasoning alone can govern the calibrated inspection stack.

\section{Experiments} \label{sec: experiments}

\subsection{Full-System Inspection Performance}
\label{sec:full-system-results}

This experiment evaluates whether FRAME improves end-to-end reliability and efficiency over fixed exhaustive inspection and direct VLM control when sensing and manipulation capabilities are held fixed.
We use matched specifications to prevent success through a fixed verdict. Each test unit fixes the part, instruction, required scope, and initial configuration, and is executed with both a reference-PASS specification, constructed from independently measured as-built values, and a reference-FAIL specification that changes only a task-relevant conformance boundary. All autonomous methods receive the same manifests, specifications, initial observations, calibrated measurement interfaces, frozen manipulation policy within each part setting, and action, recovery, and timeout budgets. 
An episode succeeds only when valid evidence covers every required characteristic and the requirement statuses, final verdict, report, and sorting action are correct. A pair is successful only when both its PASS-spec and FAIL-spec episodes succeed. Failure to produce a binary verdict within the common limits is recorded as unresolved and routed to manual review; a released verdict based on incomplete or incorrectly localized evidence is instead scored as claimed and may become a false reject or false accept.

\begin{figure}[t]
\centering
\includegraphics[width=\columnwidth]{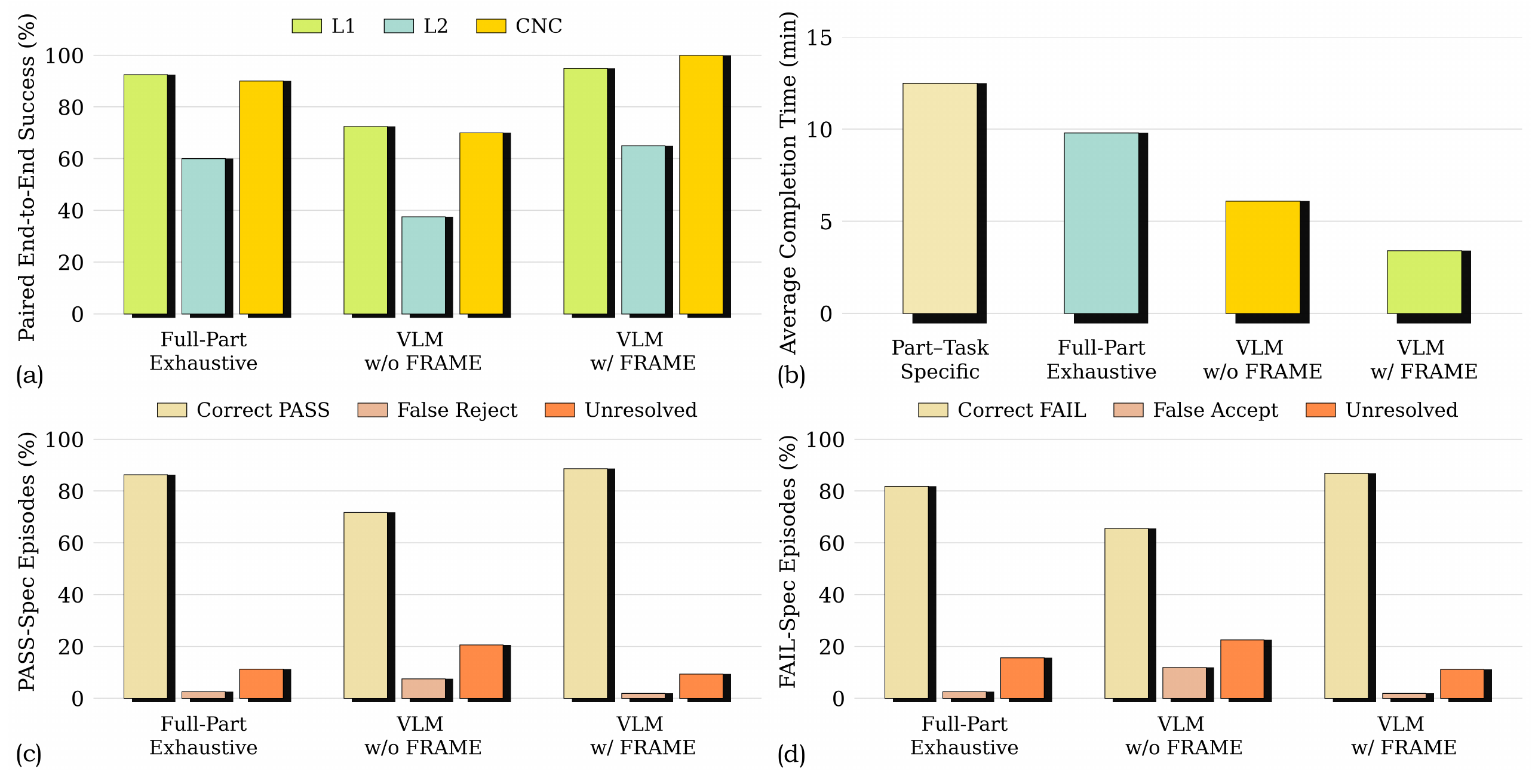}
\caption{End-to-end reliability and task completion cost.
(a) Paired end-to-end success across L1, L2, and CNC parts, where a pair is counted as successful only when both the matched PASS-spec and FAIL-spec episodes are completed correctly. (b) Mean task-to-completion time, measured from task-specific preparation to final report and sorting or unresolved termination. (c) Terminal outcomes on PASS-spec episodes, reported as correct PASS, false reject, and unresolved. (d) Terminal outcomes conditional on FAIL-spec episodes, reported as correct FAIL, conditional false accept, and unresolved. An episode is unresolved when no valid binary verdict is produced within the common execution and recovery budget and is therefore transferred to manual review.}
\label{fig:exp1}
\end{figure}

Figure~\ref{fig:exp1}(a) shows that FRAME achieves the strongest paired reliability on L1, L2, and CNC parts. The L2 result is especially informative because the autonomous methods share the same reorientation capability and differ primarily in observation, recovery, and evidence management. Exhaustive inspection accumulates failure exposure through task-irrelevant operations, whereas direct VLM control lacks authoritative correspondence and deterministic evidence gates. FRAME limits acquisition to the active scope and releases a verdict only after verified evidence completes that scope. Consistently, Fig.~\ref{fig:exp1}(c--d) shows more correct terminal outcomes and the fewest conditional false accepts among FAIL-spec episodes. Unresolved FRAME episodes occur when correspondence, reorientation, or measurement admissibility cannot be established within budget, thereby withholding an unsupported verdict rather than releasing a nonconforming part.

Figure~\ref{fig:exp1}(b) shows that FRAME also completes the workflow fastest by avoiding unnecessary surface access, scanning, and analysis and stopping when admissible coverage is complete. Exhaustive inspection processes requirements outside the requested scope, manual task-specific inspection adds interpretation and tool-configuration time, and direct VLM control may spend additional actions resolving planning or evidence uncertainty. Unresolved autonomous episodes are charged the common timeout so that early failure does not appear artificially efficient. Thus, within the supported parts and measurement vocabulary, verified task-specific evidence improves paired reliability, false-accept protection, and completion time.

\subsection{Metrological Validation}
\label{sec:measurement-qualification}

\begin{figure}[t]
\centering
\includegraphics[width=\columnwidth]{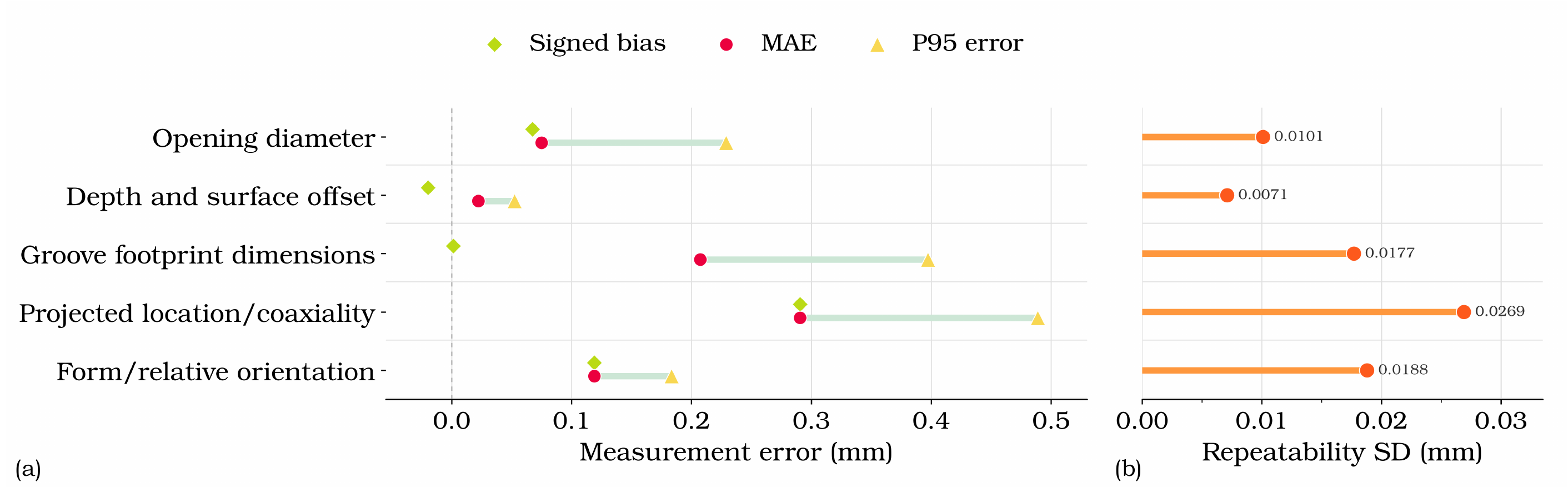}
\caption{Metrology validation across representative measurement operators. (a) Signed bias, mean absolute error (MAE), and 95th-percentile absolute error (P95) relative to reference measurements for opening diameter, depth/surface offset, groove footprint dimensions, projected location/coaxiality, and form/relative orientation. The dashed vertical line denotes zero signed bias. (b) Repeatability standard deviation over repeated measurements for the same operator groups, quantifying scan-to-scan measurement consistency.}
\label{fig:exp_measurement}
\end{figure}

The second experiment validates the measurement expert independently of task-planning and manipulation outcomes. Each physical characteristic is measured with a dedicated reference instrument to establish its as-built value. The evaluation covers every inspectable face of every benchmark object. For each face, the workpiece is placed at three admissible in-plane orientations relative to the profiler sweep direction, and three repeated scans are acquired at each orientation to quantify repeatability. Every acquisition is processed by the same deterministic pipeline used in the full-system experiments: streamed laser profiles are assembled into a height map, residual tilt is removed by robust plane fitting, height levels and feature candidates are extracted, and the corresponding datum-grounded operator produces the reported quantity. For each operator family, Fig.~\ref{fig:exp_measurement}(a) evaluates agreement with the reference instrument using signed bias, mean absolute error, and tail absolute error, whereas Fig.~\ref{fig:exp_measurement}(b) evaluates scan-to-scan repeatability using the standard deviation of repeated measurements. Repeated acquisitions characterize measurement stability for the available physical features and are not treated as additional independently manufactured specimens.

Each row on the vertical axis of Fig.~\ref{fig:exp_measurement} groups quantities that share a geometric estimator. \emph{Opening diameter} denotes diameters recovered by fitting circular contours to blind, through, or nested openings. \emph{Depth and surface offset} denotes the vertical separation between a local host surface and a recessed feature, or between adjacent surface levels. \emph{Groove footprint dimensions} denotes the in-plane length and width obtained from the oriented extent of a segmented groove. \emph{Projected location/coaxiality} denotes datum-referenced feature-center displacement or the relative offset between fitted opening centers. Finally, \emph{form/relative orientation} denotes plane-based characteristics, including flatness and parallelism, derived from fitted surface residuals or relations between fitted planes.

Depth and surface-offset estimation provides the strongest overall metrological performance. This behavior is consistent with the implementation: depth is determined primarily from a robust vertical separation between the reference surface and the feature level after tilt compensation, making it comparatively insensitive to errors in the reconstructed in-plane sweep coordinate. Diameter, groove footprint, and projected-location measurements, in contrast, depend directly on the spatial spacing and shape of successive profiles. The current acquisition code reconstructs the sweep axis from the commanded robot speed and profiler rate rather than from a robot pose synchronized to every acquired profile. Residual velocity variation and jerk of the measurement arm can therefore create nonuniform physical profile spacing that is represented as a uniform grid, distorting dimensions and feature centers along the sweep direction.

A second limitation follows from the use of a single scan direction for each reported acquisition; scan direction is varied between repeats, but multiple directions are not fused within one measurement. When a feature is sufficiently deep, the laser-triangulation geometry causes one wall to occlude part of the opposite wall or bottom in a direction-dependent manner. The implementation converts the resulting missing returns to undefined height samples and fits the reference and feature planes only from the remaining valid points. A deep feature can consequently produce an asymmetric support region that perturbs plane estimation and the boundary derived from that plane, degrading form, relative-orientation, and lateral-size measurements even when the vertical level separation remains stable.

\subsection{Active Surface Correspondence}
\label{sec:correspondence-results}

\begin{table}[t]
\centering
\caption{LOOK-based surface-correspondence performance. Both methods receive the same fixed LOOK observations; FRAME additionally resolves per-view VLM scores over legal configurations and applies the confidence–consistency gate before releasing a correspondence.}
\label{tab:surface-correspondence}
\setlength{\tabcolsep}{2pt}
\renewcommand{\arraystretch}{1.15}
\begin{tabularx}{\columnwidth}{L{0.34\columnwidth} *{4}{Y}}
\toprule
\shortstack[l]{\textbf{Method}\\\strut} & \shortstack{\textbf{Cand. acc.}\\\textbf{(\%) $\uparrow$}} & \shortstack{\textbf{Wrong rel.}\\\textbf{/ trial (\%) $\downarrow$}} & \shortstack{\textbf{Abstain}\\\textbf{(\%)}} & \shortstack{\textbf{Rel. acc.}\\\textbf{(\%) $\uparrow$}} \\
\midrule
\multicolumn{5}{l}{\emph{L1}} \\
VLM w/o FRAME & 54.8 & 45.2 & - & - \\
\textbf{VLM w/ FRAME} & \textbf{92.9} & \textbf{2.4} & 9.5 & \textbf{97.4} \\
\addlinespace[2pt]
\cmidrule(lr){1-5}
\addlinespace[1pt]
\multicolumn{5}{l}{\emph{L2}} \\
VLM w/o FRAME & 52.4 & 47.6 & - & - \\
\textbf{VLM w/ FRAME} & \textbf{82.1} & \textbf{3.6} & 19.0 & \textbf{95.6} \\
\bottomrule
\end{tabularx}
\end{table}

We evaluate LOOK based physical to specification correspondence independently of scanning and measurement by comparing FRAME with \emph{Direct VLM Judgment}. Both methods receive exactly the same fixed LOOK observations for each geometry family. When a configuration requires one LOOK, the baseline receives that single image. When it requires multiple LOOKs, all selected images are provided together, and the VLM is asked to directly return the physical to specification region assignment. The prediction is released without legal configuration resolution or a confidence and consistency gate. FRAME instead uses the VLM only to produce region matching scores for each view, resolves those scores over the finite set of legal configurations, and releases the resulting correspondence only if its score, top two margin, and anchor and topology checks all pass. The fixed LOOK set is executed once in both methods. A failed FRAME gate produces abstention and never triggers an additional LOOK.

\begin{figure}[t]
\centering
\includegraphics[width=\columnwidth]{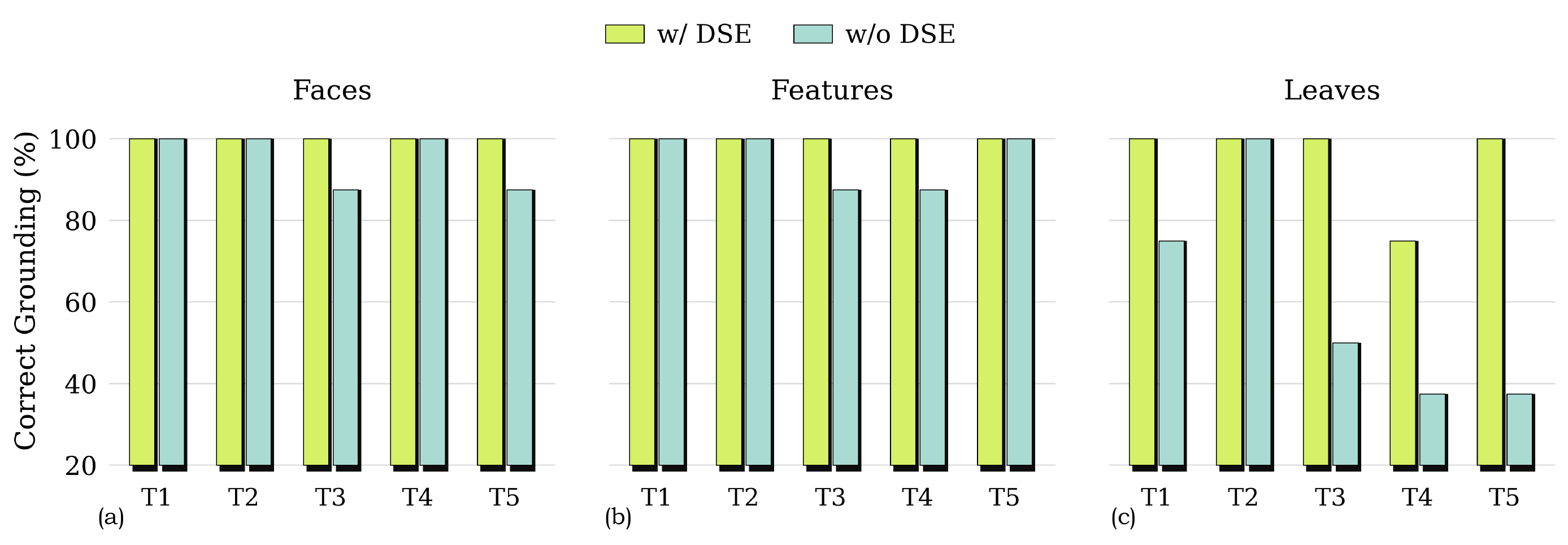}
\caption{Scope risk evaluation with and without deterministic scope expansion (DSE). Correct grounding rates are reported at three levels of the specification hierarchy: (a) required faces, (b) corresponding features, and (c) leaf requirements, across task types T1–T5. }
\label{fig:exp4a}
\end{figure}

Table~\ref{tab:surface-correspondence} reports four complementary metrics. \emph{Candidate accuracy} is the fraction of trials whose candidate correspondence is correct before the gate. For FRAME, this evaluates the highest scoring legally resolved configuration. \emph{Per-trial wrong-release rate} is the fraction of all trials in which an incorrect correspondence is released and can therefore enter the evidence path. \emph{Abstention rate} is the fraction of trials for which FRAME releases no correspondence because at least one gate condition fails. \emph{Released accuracy} is the fraction of released FRAME correspondences that are correct, so it measures the reliability of released outputs rather than the accuracy of the gate as a binary classifier. Direct VLM Judgment has no gate, so its abstention rate and released accuracy are marked as not applicable. Because every baseline prediction is released, its wrong release rate equals its candidate error rate.

We report L1 and L2 separately to expose the effect of morphology and visual complexity. The benchmark contains one L1 and two L2 workpieces from each of four families: cube, stepped block, cylinder, and bracket. This gives four L1 and eight L2 workpieces. For every cube and stepped block workpiece, we test five legal configurations. For every cylinder and bracket, we test two. Each configuration is executed three times with the prescribed LOOK set, and ground truth physical to specification associations are annotated for every execution. This protocol yields 42 L1 trials and 84 L2 trials. Both methods are evaluated on paired observations from the same executions so that the comparison isolates correspondence inference and gating rather than viewpoint or acquisition variation.

The table shows that directly requesting a complete correspondence from the VLM is unreliable, particularly when the decision requires combining multiple LOOKs. A single dominant view often provides enough evidence for the VLM to identify the visible region, but input with multiple views introduces hallucination and inconsistent binding between images and specification regions. The error audit also reveals failures on fine grained geometry. The VLM may miscount repeated holes, omit a counterbore, or assign several observed physical faces to the same specification face. Such outputs can appear semantically plausible while violating the object's adjacency, opposite face, handedness, or exclusivity relations, and the unconstrained baseline has no mechanism to reject them.

FRAME largely prevents these failures by converting the VLM evidence from each view into scores over only legal configurations and withholding outputs that are weak, ambiguous, or structurally inconsistent. The resulting pattern is higher candidate accuracy, substantially fewer wrong releases, and high accuracy among released outputs, with increased abstention on the more challenging L2 workpieces. Residual errors mainly arise from observation quality rather than unconstrained correspondence generation. In particular, an oblique wrist camera viewpoint can cause the SAM3 crop to include multiple physical faces. The resulting mixed visual evidence may lower the score or margin and trigger abstention. In rare cases, it may support a coherent but incorrect legal configuration that passes the gate. 

\begin{table}[t]
\centering
\caption{Post-reorientation state admission on matched L2 executions with and without the correspondence gate. Rejected and abstained states are counted as not admitted.}
\label{tab:correspondence-gate}
\setlength{\tabcolsep}{6pt}
\renewcommand{\arraystretch}{1.15}
\footnotesize
\begin{tabular}{lcc}
\toprule
\shortstack[l]{\textbf{Method}\\\strut}
&
\shortstack{\textbf{Correct-face}\\\textbf{acceptance} $\uparrow$}
&
\shortstack{\textbf{Wrong-face}\\\textbf{admission} $\downarrow$}
\\
\midrule
w/o correspondence gate
& 15/15
& 25/25
\\
\textbf{w/ correspondence gate}
& \textbf{14/15}
& \textbf{2/25}
\\
\bottomrule
\end{tabular}
\end{table}

\begin{figure}[t]
\centering
\includegraphics[width=\columnwidth]{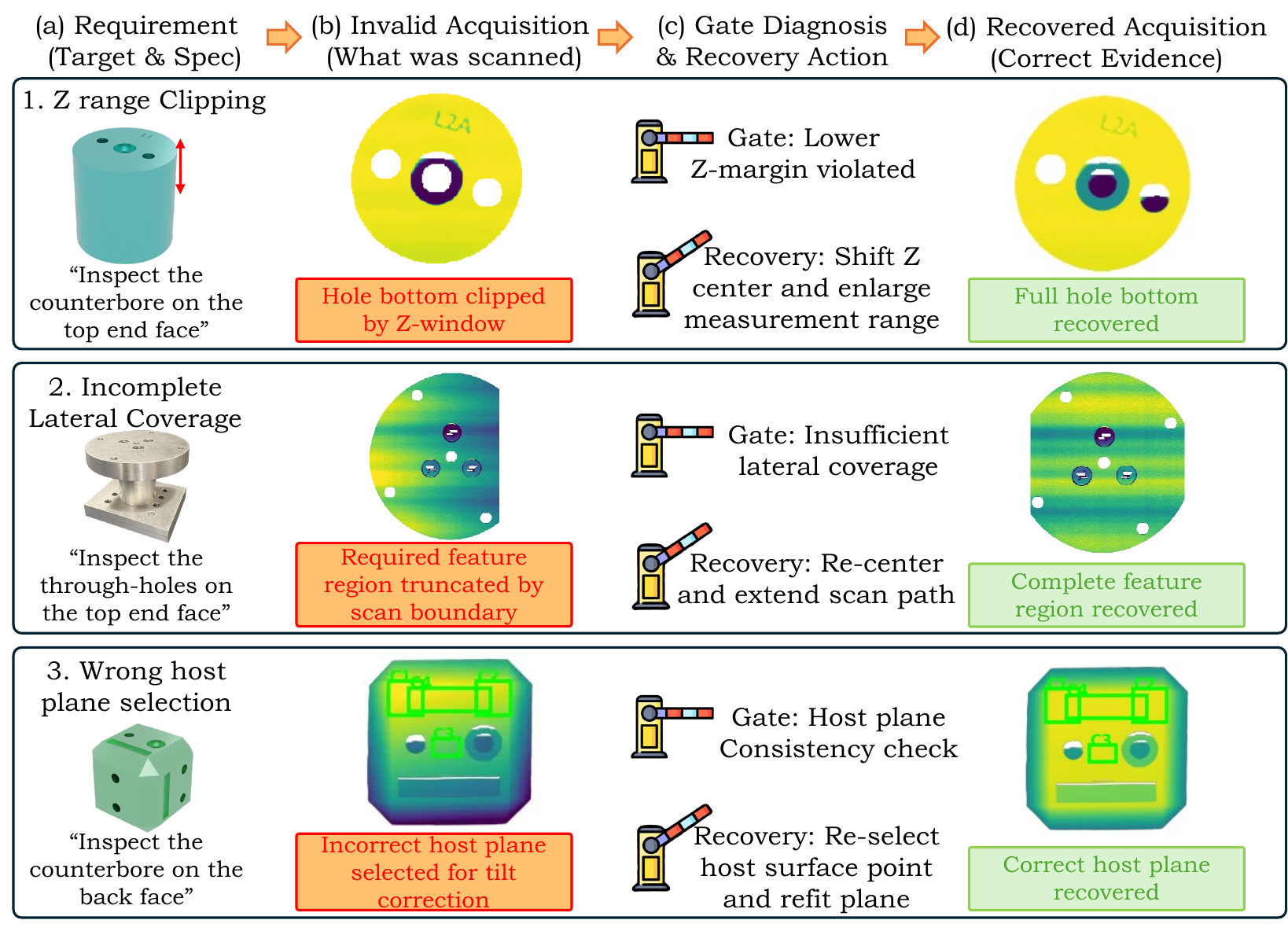}
\caption{Representative metrology failures and recovery. Three cases illustrate Z-range clipping, incomplete lateral coverage, and incorrect host-plane selection. Requirement-specific admissibility checks reject the invalid acquisition and trigger targeted recovery before the measurement is admitted as evidence.}
\label{fig:exp4c}
\end{figure}

\begin{figure*}[t]
\centering
\includegraphics[width=\textwidth]{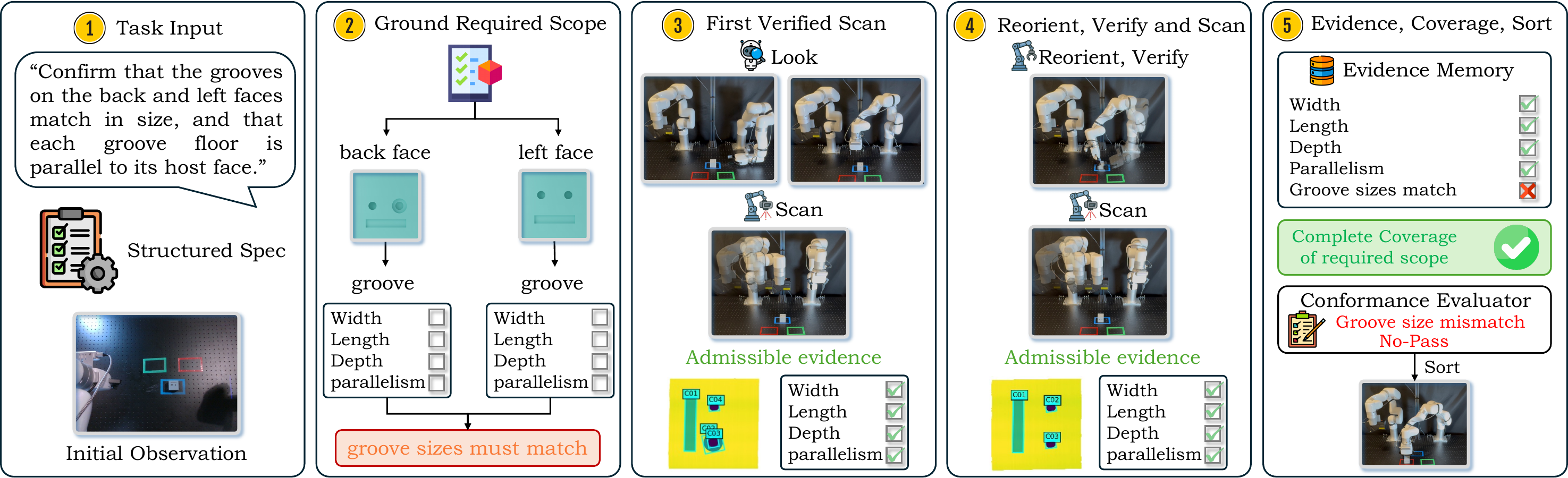}
\caption{Case study of cross-face groove inspection. Given the instruction and structured specification, FRAME expands the required scope from target faces to groove features and leaf requirements, including width, length, depth, floor parallelism, and the cross-face size relation. The system verifies the first surface before scanning, admits only valid measurements into Evidence Memory, then reorients the part, re-verifies the exposed surface, and performs the second scan. After complete evidence coverage is achieved, the deterministic conformance evaluator identifies a groove-size mismatch and issues a NO-PASS verdict before sorting.}
\label{fig:case-study}
\end{figure*}

\subsection{Risk Decomposition}

\subsubsection{Scope Risk ($\gamma$)}

Figure~\ref{fig:exp4a} evaluates whether the grounded scope contains all ground-truth faces, features, and atomic leaf requirements. DSE and its ablation perform similarly at the face and feature levels but diverge substantially at the leaf level. The dominant error is therefore incomplete enumeration rather than failure to recognize the high-level target. By expanding specification-declared dependencies, DSE prevents member measurements, cross-surface relations, and supporting requirements from being omitted, particularly for composite, relational, and functional-goal instructions.

\subsubsection{Correspondence Risk ($\delta$)}

This experiment isolates post-reorientation verification using all eight L2 workpieces and five task types, with one execution per object--task pair. Each episode begins from a random admissible pose in which the required canonical face $f_{\mathrm{exp}}$ is not scannable. After the frozen VLA attempts reorientation, a calibrated \textsc{Look} updates the physical-to-specification correspondence; terminal states are annotated independently from the calibrated view, feature registry, and video. Both variants use the same VLA executions and differ only in whether the updated correspondence must verify $f_{\mathrm{exp}}$ before measurement. Without the gate, every completed action is admitted; with it, a mismatch, rejection, or abstention triggers recovery or manual review.
Table~\ref{tab:correspondence-gate} shows that the ungated variant accepts all 15 correct-face states but also all 25 wrong-face states. The gate retains 14 of 15 correct states while reducing wrong-face admission to 2 of 25. It does not improve manipulation; instead, it detects most reorientation failures before they propagate as wrong-face evidence. Only an incorrect correspondence that passes the gate contributes to $\delta$.

\subsubsection{Metrology Risk ($\alpha$)}

We isolate metrology risk $\alpha$ by fixing the scope, correspondence, operator, and specification while perturbing only acquisition or measurement processing. Requirement-specific admissibility predicates reject plausible but invalid measurements before they enter $\mathcal{M}_t$; the failed predicate then selects a targeted recovery, after which the same operator and checks are reapplied.
Figure~\ref{fig:exp4c} illustrates this behavior for Z-range clipping, incomplete lateral coverage, and incorrect host-plane selection. The corresponding checks respectively adjust the Z window, extend and recenter the scan, or reselect support points before refitting. In each case, the recovered result must again pass admissibility, preventing acquisition or reconstruction failures from supporting conformance and reducing $\alpha$ without relying on a generic rescan.

\subsection{Case Study}

Fig.~\ref{fig:case-study} demonstrates how FRAME resolves cross-face evidence requirements. For an instruction comparing grooves on the back and left faces, the system grounds the required groove dimensions, floor parallelism, and cross-face size relation, then verifies the exposed surface before scanning. Because the first scan leaves the left-face requirements uncovered, the coverage audit triggers reorientation; a second \textsc{Look} confirms surface correspondence before the next scan is admitted. Once both sets of admissible measurements enter Evidence Memory, the deterministic evaluator identifies a groove-size mismatch and returns \textsc{No-Pass}, despite valid individual groove and parallelism measurements. Thus, reorientation serves only unresolved evidence demand, while verified evidence rather than the manipulation outcome determines the final verdict and sorting action.

\section{Conclusion} \label{sec: conclusion}

This paper introduced task-specified active metrological
inspection and a hierarchical dual-arm framework that turns
an inspection instruction and a structured specification into
complete, traceable evidence. Learned components ground the
requested task and provide physical access, while deterministic
measurement, admissibility, coverage, and conformance gates
control which evidence may authorize a release decision.
Physical experiments on controlled workpiece families and
real CNC parts show improved end-to-end reliability and com-
pletion efficiency over exhaustive execution and unconstrained
VLM control, while component tests demonstrate the value of
correspondence verification and measurement-failure rejection.
The current system remains limited to a finite specification and
measurement vocabulary, support-stable configurations, and a
single profilometry setup; moreover, motion-derived sweep
reconstruction and single-direction acquisition can degrade
lateral and form measurements for difficult geometries. The
false-accept decomposition is also an empirically qualified risk
contract rather than an unconditional deployment-wide guar-
antee. Future work will incorporate synchronized robot poses
and multi-direction scan fusion, compile broader MBD/PMI
semantics into executable inspection rules, and evaluate the
framework across more diverse geometries, materials, sensors,
and production cells.

\section*{Acknowledgments}

This work was supported under Cooperative Agreement W56HZV-21-2-0001 with the U.S. Army DEVCOM Ground Vehicle Systems Center (GVSC), through the Virtual Prototyping of Autonomy Enabled Ground Systems (VIPR-GS) program, and by the National Science Foundation, United States (Grant Nos. 2434519 and 2543603).

DISTRIBUTION STATEMENT A. Approved for public release; distribution is unlimited. OPSEC11090

Disclaimer: Reference herein to any specific commercial company, product, process, or service by trade name, trademark, manufacturer, or otherwise, does not necessarily constitute or imply its endorsement, recommendation, or favoring by the United States Government or the Department of the Army (DoA). The opinions of the authors expressed herein do not necessarily state or reflect those of the United States Government or the DoA and shall not be used for advertising or product endorsement purposes.


%





\ifCLASSOPTIONcaptionsoff
  \newpage
\fi





\bibliographystyle{IEEEtran}
\bibliography{IEEEabrv,Bibliography}

@IEEEtranBSTCTL{IEEEexample:BSTcontrol,
  CTLuse_article_number      = "yes",
  CTLuse_paper               = "yes",
  CTLuse_forced_etal         = "no",
  CTLmax_names_forced_etal   = "50",
  CTLnames_show_etal         = "50",
  CTLuse_alt_spacing         = "yes",
  CTLalt_stretch_factor      = "4",
  CTLdash_repeated_names     = "yes",
  CTLname_format_string      = "{f.~}{vv~}{ll}{, jj}",
  CTLname_latex_cmd          = "",
  CTLname_url_prefix         = "[Online]. Available:"
}

@inproceedings{liang2023code,
  title={Code as policies: Language model programs for embodied control},
  author={Liang, Jacky and Huang, Wenlong and Xia, Fei and Xu, Peng and Hausman, Karol and Ichter, Brian and Florence, Pete and Zeng, Andy},
  booktitle={2023 IEEE International conference on robotics and automation (ICRA)},
  pages={9493--9500},
  year={2023},
  organization={IEEE}
}

@inproceedings{huang2023voxposer,
  title={VoxPoser: Composable 3D Value Maps for Robotic Manipulation with Language Models},
  author={Huang, Wenlong and Wang, Chen and Zhang, Ruohan and Li, Yunzhu and Wu, Jiajun and Fei-Fei, Li},
  booktitle={Conference on Robot Learning},
  pages={540--562},
  year={2023},
  organization={PMLR}
}

@inproceedings{kim2025openvla,
  title={OpenVLA: An Open-Source Vision-Language-Action Model},
  author={Kim, Moo Jin and Pertsch, Karl and Karamcheti, Siddharth and Xiao, Ted and Balakrishna, Ashwin and Nair, Suraj and Rafailov, Rafael and Foster, Ethan P and Sanketi, Pannag R and Vuong, Quan and others},
  booktitle={Conference on Robot Learning},
  pages={2679--2713},
  year={2025},
  organization={PMLR}
}

@article{zhang2026harness,
  title={Harness VLA: Steering Frozen VLAs into Reliable Manipulation Primitives via Memory-Guided Agents},
  author={Zhang, Yixian and Zhang, Huanming and Gao, Feng and Li, Xiao and Liu, Zhihao and Zhu, Chunyang and Qiu, Jiaxing and Yan, Yuchen and Liu, Jiyuan and Tang, Wenhao and others},
  journal={arXiv preprint arXiv:2607.08448},
  year={2026}
}

@manual{asme_y145_2018,
  author       = {{ASME}},
  title        = {Dimensioning and Tolerancing},
  organization = {The American Society of Mechanical Engineers},
  address      = {New York, NY, USA},
  year         = {2018},
  note         = {ASME Y14.5-2018 (R2024)}
}

@article{black2026real,
  title={Real-time execution of action chunking flow policies},
  author={Black, Kevin and Galliker, Manuel and Levine, Sergey},
  journal={Advances in Neural Information Processing Systems},
  volume={38},
  pages={33383--33407},
  year={2026}
}

@article{hu2026matters,
  title={What Matters in Orchestrating Robot Policies: A Systematic Study of Hierarchical VLA Agents},
  author={Hu, Jiaheng and Shridhar, Mohit and Lu, Caden and Shah, Dhruv and Chiang, Hao-Tien Lewis and Tan, Jie and Xie, Annie},
  journal={arXiv preprint arXiv:2606.10267},
  year={2026}
}

@article{Qwen3-VL,
      title={Qwen3-VL Technical Report}, 
      author={Shuai Bai and Yuxuan Cai and Ruizhe Chen and Keqin Chen and Xionghui Chen and Zesen Cheng and Lianghao Deng and Wei Ding and Chang Gao and Chunjiang Ge and Wenbin Ge and Zhifang Guo and Qidong Huang and Jie Huang and Fei Huang and Binyuan Hui and Shutong Jiang and Zhaohai Li and Mingsheng Li and Mei Li and Kaixin Li and Zicheng Lin and Junyang Lin and Xuejing Liu and Jiawei Liu and Chenglong Liu and Yang Liu and Dayiheng Liu and Shixuan Liu and Dunjie Lu and Ruilin Luo and Chenxu Lv and Rui Men and Lingchen Meng and Xuancheng Ren and Xingzhang Ren and Sibo Song and Yuchong Sun and Jun Tang and Jianhong Tu and Jianqiang Wan and Peng Wang and Pengfei Wang and Qiuyue Wang and Yuxuan Wang and Tianbao Xie and Yiheng Xu and Haiyang Xu and Jin Xu and Zhibo Yang and Mingkun Yang and Jianxin Yang and An Yang and Bowen Yu and Fei Zhang and Hang Zhang and Xi Zhang and Bo Zheng and Humen Zhong and Jingren Zhou and Fan Zhou and Jing Zhou and Yuanzhi Zhu and Ke Zhu},
	  journal={arXiv preprint arXiv:2511.21631},
      year={2025}
}

@misc{intelligence2025pi05visionlanguageactionmodelopenworld,
      title={$\pi_{0.5}$: a Vision-Language-Action Model with Open-World Generalization}, 
      author={Physical Intelligence and Kevin Black and Noah Brown and James Darpinian and Karan Dhabalia and Danny Driess and Adnan Esmail and Michael Equi and Chelsea Finn and Niccolo Fusai and Manuel Y. Galliker and Dibya Ghosh and Lachy Groom and Karol Hausman and Brian Ichter and Szymon Jakubczak and Tim Jones and Liyiming Ke and Devin LeBlanc and Sergey Levine and Adrian Li-Bell and Mohith Mothukuri and Suraj Nair and Karl Pertsch and Allen Z. Ren and Lucy Xiaoyang Shi and Laura Smith and Jost Tobias Springenberg and Kyle Stachowicz and James Tanner and Quan Vuong and Homer Walke and Anna Walling and Haohuan Wang and Lili Yu and Ury Zhilinsky},
      year={2025},
      eprint={2504.16054},
      archivePrefix={arXiv},
      primaryClass={cs.LG},
      url={https://arxiv.org/abs/2504.16054}, 
}

@article{chen2026gap,
  title={GaP: A Graph-as-Policy Multi-Agent Self-Learning Harness For Variational Automation Tasks},
  author={Chen, Kaiyuan and Xie, Shuangyu and Fu, Letian and Yu, Justin and Pacini, William and Bajamahal, Sandeep and Kim, Hudson and Drake, Jaimyn and Kim, Daehwa and Xue, Haoru and others},
  journal={arXiv preprint arXiv:2607.05369},
  year={2026}
}

@article{brohan2023rt,
  title={Rt-2: Vision-language-action models transfer web knowledge to robotic control},
  author={Brohan, Anthony and Brown, Noah and Carbajal, Justice and Chebotar, Yevgen and Chen, Xi and Choromanski, Krzysztof and Ding, Tianli and Driess, Danny and Dubey, Avinava and Finn, Chelsea and others},
  journal={arXiv preprint arXiv:2307.15818},
  year={2023}
}

@misc{black2026pi0visionlanguageactionflowmodel,
      title={$\pi_0$: A Vision-Language-Action Flow Model for General Robot Control}, 
      author={Kevin Black and Noah Brown and Danny Driess and Adnan Esmail and Michael Equi and Chelsea Finn and Niccolo Fusai and Lachy Groom and Karol Hausman and Brian Ichter and Szymon Jakubczak and Tim Jones and Liyiming Ke and Sergey Levine and Adrian Li-Bell and Mohith Mothukuri and Suraj Nair and Karl Pertsch and Lucy Xiaoyang Shi and James Tanner and Quan Vuong and Anna Walling and Haohuan Wang and Ury Zhilinsky},
      year={2026},
      eprint={2410.24164},
      archivePrefix={arXiv},
      primaryClass={cs.LG},
      url={https://arxiv.org/abs/2410.24164}, 
}

@article{ye2026world,
  title={World action models are zero-shot policies},
  author={Ye, Seonghyeon and Ge, Yunhao and Zheng, Kaiyuan and Gao, Shenyuan and Yu, Sihyun and Kurian, George and Indupuru, Suneel and Tan, You Liang and Zhu, Chuning and Xiang, Jiannan and others},
  journal={arXiv preprint arXiv:2602.15922},
  year={2026}
}

@article{fan2025long,
  title={Long-vla: Unleashing long-horizon capability of vision language action model for robot manipulation},
  author={Fan, Yiguo and Ding, Pengxiang and Bai, Shuanghao and Tong, Xinyang and Zhu, Yuyang and Lu, Hongchao and Dai, Fengqi and Zhao, Wei and Liu, Yang and Huang, Siteng and others},
  journal={arXiv preprint arXiv:2508.19958},
  year={2025}
}

@article{liu2026palm,
  title={Palm: Progress-aware policy learning via affordance reasoning for long-horizon robotic manipulation},
  author={Liu, Yuanzhe and Zhu, Jingyuan and Mo, Yuchen and Li, Gen and Cao, Xu and Jin, Jin and Shen, Yifan and Li, Zhengyuan and Yu, Tianjiao and Yuan, Wenzhen and others},
  journal={arXiv preprint arXiv:2601.07060},
  year={2026}
}

@article{yang2024swe,
  title={Swe-agent: Agent-computer interfaces enable automated software engineering},
  author={Yang, John and Jimenez, Carlos and Wettig, Alexander and Lieret, Kilian and Yao, Shunyu and Narasimhan, Karthik and Press, Ofir},
  journal={Advances in Neural Information Processing Systems},
  volume={37},
  pages={50528--50652},
  year={2024}
}

@article{zhong2026ai,
  title={Ai harness engineering: A runtime substrate for foundation-model software agents},
  author={Zhong, Hailin and Zhu, Shengxin},
  journal={arXiv preprint arXiv:2605.13357},
  year={2026}
}

@article{lee2026harness,
  title={Harness Engineering for Physical AI: Robot Middleware Is the Harness Layer},
  author={Lee, Sanghoon and Chae, Jiyeong and Park, Kyung-Joon},
  journal={arXiv preprint arXiv:2606.09416},
  year={2026}
}

@article{lin2001cad,
  title={CAD-based CMM dimensional inspection path planning--a generic algorithm},
  author={Lin, Yueh-Jaw and Mahabaleshwarkar, Rahul and Massina, Elena},
  journal={Robotica},
  volume={19},
  number={2},
  pages={137--148},
  year={2001},
  publisher={Cambridge University Press}
}

@article{phan2018path,
  title={Path planning of a laser-scanner with the control of overlap for 3d part inspection},
  author={Phan, Nguyen Duy Minh and Quinsat, Yann and Lavernhe, Sylvain and Lartigue, Claire},
  journal={Procedia Cirp},
  volume={67},
  pages={392--397},
  year={2018},
  publisher={Elsevier}
}

@article{vlaeyen2022uncertainty,
  title={Uncertainty-based autonomous path planning for laser line scanners},
  author={Vlaeyen, Michiel and Haitjema, Han and Dewulf, Wim},
  journal={Metrology},
  volume={2},
  number={4},
  pages={479--494},
  year={2022},
  publisher={MDPI}
}

@inproceedings{chen2024gennbv,
  title={Gennbv: Generalizable next-best-view policy for active 3d reconstruction},
  author={Chen, Xiao and Li, Quanyi and Wang, Tai and Xue, Tianfan and Pang, Jiangmiao},
  booktitle={2024 IEEE/CVF Conference on Computer Vision and Pattern Recognition (CVPR)},
  pages={16436--16445},
  year={2024},
  organization={IEEE}
}

@article{chen2025scanbot,
  title={ScanBot: A Benchmark for Precision Robotic Surface Scanning with Industrial Laser Profilers},
  author={Chen, Zhiling and Zhang, Yang and Piran, Fardin Jalil and Zhou, Qianyu and Tang, Jiong and Imani, Farhad},
  journal={arXiv preprint arXiv:2505.17295},
  year={2025}
}

@inproceedings{bergmann2019mvtec,
  title={MVTec AD—A comprehensive real-world dataset for unsupervised anomaly detection},
  author={Bergmann, Paul and Fauser, Michael and Sattlegger, David and Steger, Carsten},
  booktitle={2019 IEEE/CVF Conference on Computer Vision and Pattern Recognition (CVPR)},
  pages={9584--9592},
  year={2019},
  organization={IEEE}
}

@inproceedings{roth2022towards,
  title={Towards total recall in industrial anomaly detection},
  author={Roth, Karsten and Pemula, Latha and Zepeda, Joaquin and Sch{\"o}lkopf, Bernhard and Brox, Thomas and Gehler, Peter},
  booktitle={2022 IEEE/CVF Conference on Computer Vision and Pattern Recognition (CVPR)},
  pages={14298--14308},
  year={2022},
  organization={IEEE}
}

@inproceedings{gu2024anomalygpt,
  title={Anomalygpt: Detecting industrial anomalies using large vision-language models},
  author={Gu, Zhaopeng and Zhu, Bingke and Zhu, Guibo and Chen, Yingying and Tang, Ming and Wang, Jinqiao},
  booktitle={Proceedings of the AAAI conference on artificial intelligence},
  volume={38},
  number={3},
  pages={1932--1940},
  year={2024}
}

@article{brohan2022rt,
  title={Rt-1: Robotics transformer for real-world control at scale},
  author={Brohan, Anthony and Brown, Noah and Carbajal, Justice and Chebotar, Yevgen and Dabis, Joseph and Finn, Chelsea and Gopalakrishnan, Keerthana and Hausman, Karol and Herzog, Alex and Hsu, Jasmine and others},
  journal={arXiv preprint arXiv:2212.06817},
  year={2022}
}

@article{bousmalis2023robocat,
  title={Robocat: A self-improving generalist agent for robotic manipulation},
  author={Bousmalis, Konstantinos and Vezzani, Giulia and Rao, Dushyant and Devin, Coline and Lee, Alex X and Bauz{\'a}, Maria and Davchev, Todor and Zhou, Yuxiang and Gupta, Agrim and Raju, Akhil and others},
  journal={arXiv preprint arXiv:2306.11706},
  year={2023}
}

@article{team2024octo,
  title={Octo: An open-source generalist robot policy},
  author={Team, Octo Model and Ghosh, Dibya and Walke, Homer and Pertsch, Karl and Black, Kevin and Mees, Oier and Dasari, Sudeep and Hejna, Joey and Kreiman, Tobias and Xu, Charles and others},
  journal={arXiv preprint arXiv:2405.12213},
  year={2024}
}

@inproceedings{huang2022language,
  title={Language models as zero-shot planners: Extracting actionable knowledge for embodied agents},
  author={Huang, Wenlong and Abbeel, Pieter and Pathak, Deepak and Mordatch, Igor},
  booktitle={International conference on machine learning},
  pages={9118--9147},
  year={2022},
  organization={PMLR}
}

@inproceedings{brohan2023can,
  title={Do as i can, not as i say: Grounding language in robotic affordances},
  author={Brohan, Anthony and Chebotar, Yevgen and Finn, Chelsea and Hausman, Karol and Herzog, Alexander and Ho, Daniel and Ibarz, Julian and Irpan, Alex and Jang, Eric and Julian, Ryan and others},
  booktitle={Conference on robot learning},
  pages={287--318},
  year={2023},
  organization={Pmlr}
}

@article{huang2022inner,
  title={Inner monologue: Embodied reasoning through planning with language models},
  author={Huang, Wenlong and Xia, Fei and Xiao, Ted and Chan, Harris and Liang, Jacky and Florence, Pete and Zeng, Andy and Tompson, Jonathan and Mordatch, Igor and Chebotar, Yevgen and others},
  journal={arXiv preprint arXiv:2207.05608},
  year={2022}
}

@article{yao2022react,
  title={React: Synergizing reasoning and acting in language models},
  author={Yao, Shunyu and Zhao, Jeffrey and Yu, Dian and Du, Nan and Shafran, Izhak and Narasimhan, Karthik and Cao, Yuan},
  journal={arXiv preprint arXiv:2210.03629},
  year={2022}
}

@article{shinn2023reflexion,
  title={Reflexion: Language agents with verbal reinforcement learning},
  author={Shinn, Noah and Cassano, Federico and Gopinath, Ashwin and Narasimhan, Karthik and Yao, Shunyu},
  journal={Advances in neural information processing systems},
  volume={36},
  pages={8634--8652},
  year={2023}
}

@article{liu2024demonstrating,
  title={Demonstrating ok-robot: What really matters in integrating open-knowledge models for robotics},
  author={Liu, Peiqi and Orru, Yaswanth and Vakil, Jay and Paxton, Chris and Shafiullah, Nur and Pinto, Lerrel},
  journal={Robotics: Science and Systems XX},
  year={2024},
  publisher={Robotics: Science and Systems Foundation}
}

@article{huang2024rekep,
  title={Rekep: Spatio-temporal reasoning of relational keypoint constraints for robotic manipulation},
  author={Huang, Wenlong and Wang, Chen and Li, Yunzhu and Zhang, Ruohan and Fei-Fei, Li},
  journal={arXiv preprint arXiv:2409.01652},
  year={2024}
}

@article{li2026roboclaw,
  title={Roboclaw: An agentic framework for scalable long-horizon robotic tasks},
  author={Li, Ruiying and Zhou, Yunlang and Zhu, YuYao and Chen, Kylin and Wang, Jingyuan and Wang, Sukai and Hu, Kongtao and Yu, Minhui and Jiang, Bowen and Su, Zhan and others},
  journal={arXiv preprint arXiv:2603.11558},
  year={2026}
}

@misc{hu2021loralowrankadaptationlarge,
      title={LoRA: Low-Rank Adaptation of Large Language Models}, 
      author={Edward J. Hu and Yelong Shen and Phillip Wallis and Zeyuan Allen-Zhu and Yuanzhi Li and Shean Wang and Lu Wang and Weizhu Chen},
      year={2021},
      eprint={2106.09685},
      archivePrefix={arXiv},
      primaryClass={cs.CL},
      url={https://arxiv.org/abs/2106.09685}, 
}

@inproceedings{carion2026sam,
  title={Sam 3: Segment anything with concepts},
  author={Carion, Nicolas and Gustafson, Laura and Hu, Yuan-Ting and Debnath, Shoubhik and Hu, Ronghang and Suris Coll-Vinent, Didac and Ryali, Chaitanya and Alwala, Kalyan Vasudev and Khedr, Haitham and Huang, Andrew and others},
  booktitle={International Conference on Learning Representations},
  volume={2026},
  pages={138846--138923},
  year={2026}
}

@article{yuan2026fast,
  title={Fast-wam: Do world action models need test-time future imagination?},
  author={Yuan, Tianyuan and Dong, Zibin and Liu, Yicheng and Zhao, Hang},
  journal={arXiv preprint arXiv:2603.16666},
  year={2026}
}

@article{ma2026dit4dit,
  title={Dit4dit: Jointly modeling video dynamics and actions for generalizable robot control},
  author={Ma, Teli and Zheng, Jia and Wang, Zifan and Jiang, Chunli and Cui, Andy and Liang, Junwei and Yang, Shuo},
  journal={arXiv preprint arXiv:2603.10448},
  year={2026}
}

@article{wang2026trust,
  title={When to trust imagination: Adaptive action execution for world action models},
  author={Wang, Rui and Zhang, Yue and Lin, Jiehong and Luo, Kuncheng and Wang, Jianan and Wang, Zhongrui and Qi, Xiaojuan},
  journal={arXiv preprint arXiv:2605.06222},
  year={2026}
}

@misc{thea2026,
  title  = {Towards the Harness of Embodied Agents},
  author = {Qi Wang and Tianyi Wang and Chengyang Li and Shikun Ban and Yurun Chen and Yizhong Ge and Jason Qin and Chengtai Li and Wentao Zhu},
  year   = {2026},
  note   = {Technical Report},
}

\vfill


\end{document}